\documentclass[a4paper,man, longtable, floatsintext,man]{apa7}
\usepackage{graphicx} 
\usepackage{multirow}
\usepackage{array}
\usepackage{booktabs}
\usepackage{makecell}
\usepackage{tabularx}
\usepackage{threeparttable} 
\usepackage{rotating}
\usepackage{booktabs}
\usepackage{tabularx}
\usepackage{siunitx}
\usepackage{pdflscape}
\usepackage{graphicx}
\usepackage{multirow}
\usepackage{booktabs}
\usepackage{pdflscape}
\usepackage{graphicx}
\usepackage{multirow}
\usepackage{booktabs}
\usepackage{multicol}
\usepackage{array}
\usepackage{longtable}
\usepackage{threeparttablex}
\usepackage{lipsum}
\usepackage{setspace}
\usepackage{afterpage}
\usepackage{adjustbox}
\usepackage{float}
\usepackage{enumitem}
\usepackage[table]{xcolor}
\usepackage{amsmath, amssymb, amsfonts}
\usepackage{hyperref}
\usepackage{lmodern}
\usepackage[T1]{fontenc}

\usepackage{csquotes}
\usepackage[style=apa,sortcites=true,sorting=nyt,backend=biber]{biblatex}
\DeclareLanguageMapping{american}{american-apa}
\usepackage{float}

\definecolor{LightShade}{gray}{0.9} 
\definecolor{DarkShade}{gray}{0.7}  

\newcommand{\shadefirst}[1]{\colorbox{gray!20}{\strut #1}}   
\newcommand{\shadesecond}[1]{\colorbox{gray!20}{\strut #1}}  

\shorttitle{Designing AI-Agents with Personalities}

\title{Designing AI-Agents with Personalities: A Psychometric Approach}

\authorsnames[1 2, 4, 5, 2 3]{Muhua Huang, Xijuan Zhang, Christopher Soto, James Evans}

\authorsaffiliations{
    {Stanford University},                       
    {University of Chicago Knowledge Lab},       
    {Chicago Center for Computational Social Science, Chicago}, 
    {York University},                           
    {Colby College}                              
}
\authornote{\textbf{This article has been conditionally accepted for publication in the \textit{Personality Science} journal following peer review.} This is a pre-copyedited, author-produced version of the article. 

Correspondence concerning this article should be addressed to Muhua Huang, Stanford Graduate School of Business.
Contact: muhua@stanford.edu.
}

\abstract{We introduce a methodology for assigning quantifiable and psychometrically validated personalities to AI-Agents using the Big Five framework. Across three studies, we evaluate its feasibility and limitations. In Study 1, we show that large language models (LLMs) capture semantic similarities among Big Five measures, providing a basis for personality assignment. In Study 2, we create AI-Agents using prompts designed based on the Big Five Inventory-2 (BFI-2) in different format, and find that AI-Agents powered by new models align more closely with human responses on the Mini-Markers test, although the finer pattern of results (e.g., factor loading patterns) were sometimes inconsistent. In Study 3, we validate our AI-Agents on risk-taking and moral dilemma vignettes, finding that models prompted with the BFI-2-Expanded format most closely reproduce human personality-decision associations, while safety-aligned models generally inflate ‘moral’ ratings. Overall, our results show that AI-Agents align with humans in correlations between input Big Five traits and output responses and may serve as useful tools for preliminary research. Nevertheless, discrepancies in finer response patterns indicate that AI-Agents cannot (yet) fully substitute for human participants in precision or high-stakes projects.
}

\keywords{Artificial Intelligence, AI Agents, Large Language Model, Simulation, Big Five Personalities, Psychometrics}

\begin{document}

\maketitle

\normalsize

\section{Introduction}
The emergence of large language models (LLMs) has revolutionized our approach to simulating human-like behaviors and communication. LLMs are increasingly deployed across diverse research fields to mimic human behaviors \parencite{xi_rise_2023, xu_ai_2024}. In psychology, LLMs are used to study cognitive processes, measure personality, and understand psychological constructs \parencite{hagendorff_human_like_2023, jiang_personallm_2024, rathje2024gpt, binz2025foundation}. Sociologists employ these models to explore social bias and behavior \parencite{lucy_gender_2021, zhang2025evolvingcollectivecognitionhumanagent, park_generative_2023}, while economists and political scientists utilize them to analyze economic processes and political leanings \parencite{bang-etal-2024-measuring, hartmann2023political}. These broad applications underscore the growing interconnection between artificial intelligence and the social and behavioral sciences.

Traditional human subjects research methods, while standard in social and behavioral studies, face significant challenges including ethical constraints, logistical hurdles, and resource limitations \parencite{salganik_bit_2019, demszky_using_2023}. Recent advancements in LLMs offer valuable complementary tools to address these challenges \parencite{agnew_illusion_2024, demszky_using_2023}. Researchers can use LLMs to create AI-Agents, which can mimic responses from human participants, reducing logistical and financial burdens \parencite{wang2021want}. These AI-Agents operate continuously, simulate diverse demographic responses, and can be deployed in scenarios that might pose ethical risks to human participants \parencite{argyle_out_2023, aher_using_2023, bai2022constitutional}.

The integration of AI-Agents in social and behavioral science research serves multiple practical purposes, complementing rather than replacing human participants. They can be used for pilot testing studies, allowing researchers to refine experimental designs and identify potential issues before investing in full-scale human participant studies. Additionally, AI-Agents can independently replicate findings from human-subjects data, enhancing the robustness and generalizability of research outcomes \parencite{kozlowski2024silicosociologyforecastingcovid19, zhang2025evolvingcollectivecognitionhumanagent}. This approach provides a complementary method for validation and exploration, contributing to the overall rigor and efficiency of social and behavioral science research. By offering a novel approach to corroborate and extend findings obtained through traditional human participant studies, AI-Agents represent an advancement in social science research methodology, enhancing inclusivity and efficiency while preserving the essential depth of human-centered understanding critical in the field.

\subsection{Problems with Previous Approaches for Assigning Personas to AI-Agents}
Among the myriad traits that can be incorporated into AI-Agent personas, personalities stand out as among the most intuitive characteristics to simulate. Personalities encapsulate a spectrum of human behaviors and tendencies crucial for the prediction of life outcomes ranging from academic and career to health and socioeconomic outcomes \parencite{soto_how_2019, stewart2022finer, soldz1999big} to a multitude of interaction-based phenomena \parencite{furnham1999personality, tapus2014towards}. Despite this potential and the capacity of AI-Agents to advance personality research, existing methodologies for implementing personality in AI-Agents remain limited, typically falling into three distinct approaches.

First, personas are assigned with simple personality adjectives in prompts. For example, \textcite{jiang_personallm_2024} assigned personalities to an AI-Agent by telling it ``You are a character who is introverted, antagonistic, conscientious, emotionally stable, and open to experience'' This type of approach holds a binary view, with either high-low or presence-absence of a trait, which contradicts established empirical evidence that personality traits exist on continuous spectra \parencite{asendorpf2006typeness, marcus2006antisocial, zrari2024assessing}.

Second, personas may be assigned through demographic descriptions and personal preferences \parencite{park_generative_2023, serapio2023personality, bai2025irotehumanliketraitselicitation}. For instance, \textcite{park_generative_2023} used narrative descriptions such as ``John Lin is a pharmacy shopkeeper at the Willow Market and Pharmacy who loves to help people.'' This approach relies heavily on stereotypical information extrapolated from the LLM's training corpora, where names and occupations may trigger implicit assumptions (e.g., associating "John Lin" with Asian ethnicity and pharmaceutical expertise). This approach, however, poses difficulties in evaluating the full scope of stereotypical implications (i.e., potential confounds) and fails to provide the granularity and precision necessary for social and behavioral science studies, particularly when the degree of trait expression is crucial for creating AI-Agents with specific personality profiles.


Third, personas may be embedded at the parameter level through methods like fine-tuning or direct weight editing. In one approach, researchers fine-tune an LLM on an individual-specific or group-specific text corpus, producing a distinct model persona for each dataset \parencite{tan2025democratizinglargelanguagemodels, liu2024llm}. Alternatively, knowledge editing techniques modify the model’s internal weights or latent representations to encode desired traits. For example, recent work identifies “persona vectors”, which are the specific directions in a model’s activation space associated with character traits \parencite{chen2025personavectorsmonitoringcontrolling,kim2025linear}. Applying such a vector inside the model can induce a target personality (e.g. making the model more sycophantic) without full retraining. While these parameter-level interventions alter behavior more fundamentally than in-context prompts, they still lack fine-grained control over how traits manifest. Moreover, locating and adjusting the correct parameters or latent directions for a given persona is technically challenging. The process often involves complex optimization and substantial computational resources, demanding deep expertise and making it less accessible to researchers without specialized backgrounds or institutional support.

These approaches offer a superficial semblance of personality but suffer from critical limitations. Relying on stereotypes underlying traits (e.g., naive) or roles (e.g., policing officer) does not provide the precise control necessary for rigorous social and behavioral research and does not reflect the continuity and complexity inherent in human personalities \parencite{cummings_introduction_2019}.

\subsection{Psychometric Approach for Assigning LLM-Agents Persona}
In this article, we propose creating AI-Agents with personalities through a psychometric approach by leveraging Big Five Personality theory. Historically, Big Five personality theory was developed based on the \textit{Lexical Hypothesis}, which states that personality characteristics fundamental to humans have become a part of human language, and the most important characteristics are encoded by a single word \parencite{caprara2000personality}. Based on the Lexical Hypothesis, scholars selected personality-related words from the English dictionary \parencite{allport1936trait}, refined the list of words \parencite{norman1963toward, norman19672800, wiggins1979psychological}, had participants rate themselves against those descriptors, applied factor analysis to reduce the dimensionality of the data, and eventually identified the \emph{Big Five Personality traits} \parencite{mccrae1994openness, goldberg2013alternative, fiske1949consistency, peabody1989some}: Openness (O), Conscientiousness (C), Extraversion (E), Agreeableness (A) and Neuroticism (N). 

Over the past decades, many psychometric scales\footnote{The terms ``scale'' and ``test'' are used interchangeably to refer to psychometric instruments for assessing personality traits.} measuring the Big Five traits have been developed, notably the Big Five Inventory \parencite[BFI;][]{john_big_1991}, Big Five Inventory-2 \parencite[BFI2;][]{soto_next_2017}, Mini-Markers \parencite{saucier_mini-markers_1994}, the Big Five Aspects Scale \parencite[BFAS;][]{deyoung_between_2007} and the NEO Five-Factor Inventory \parencite[NEO-FFI;][]{costa_neo_1989}. Despite having different items, these psychological measurements demonstrated high reliability, convergent validity and predictive validity in samples from diverse backgrounds \parencite{mccrae1997personality, mccrae2009physics, poortinga2002cross}. Specifically,  these psychological measurements of the Big Five can significantly predict life outcomes, such as educational achievement, socioeconomic status, health, interpersonal relationships with peers and family, and more, indicating that personality traits can indeed forecast a range of individual, interpersonal and societal behaviors effectively \parencite{soto_how_2019, ozer2006personality, roberts2007power}. 

Given that the Big Five traits and LLMs are both developed based on natural language (English), using the Big Five framework to create AI-Agents with personality is theoretically intuitive and coherent. Therefore, in this article,  we propose a new prompt engineering method that leverages items and response options from a popular Big Five scale to assign personalities to AI-Agents. 
Given the high reliability and validity of the Big Five and the rigorous research on the Big Five theory over the decades,  AI-Agents created with Big Five prompts will have personalities that are more realistic, nuanced, and fine-grained, reflecting the distribution of personality types and variability in a given population. 

Bringing psychometric principles to design and evaluate has practical benefits \parencite{wang_evaluating_2023, ye2025largelanguagemodelpsychometrics}. Integrating well-established psychometric tests (i.e., high reliability and validity) into the design of AI-Agents could ensure that the personality traits assigned to AI-Agents are stable and reflective of their programmed characteristics, enhancing the credibility and robustness of findings in applications where the consistency of AI-Agents performance is critical. In addition, as psychometric tests are designed to discern different levels of expression related to psychological traits, we can reverse engineer that variability into AI-Agents to create diversity, which is necessary for simulating real-world settings. Additionally, free from external confounds, this approach allows for precise and fine-grained manipulation over the psychological constructs of interest, allowing nuanced understanding and refinement of AI-Agents' personalities for tailored applications.

In summary, leveraging the Big Five personality test to design prompts for personality assignment of AI-Agents presents an opportunity to move beyond stereotypes and create AI-Agents with psychometrically valid traits. These traits can be quantified and controlled, are realistic and predictive of real human behavior, and are thus better suited for large-scale deployment in psychology and social science research.

\subsection{Objectives of the Present Research}

The main goal of the current research is to develop and validate AI-Agents with personalities created using psychometrically sound Big Five personality measures. To achieve this goal, we conducted a series of three interconnected studies.  In Study 1, we aim to establish a foundational understanding of personality constructs and measurements using a modern LLM technique called embedding, which converts natural language into numeric vectors based on semantic relatedness between words, phrases and sentences. Using this technique, we examine how personality-related constructs are semantically represented in LLMs.

In Study 2, we examine how to use prompt engineering to create AI-Agents with personality using a popular and validated personality measure called the Big-Five Inventory-2 (BFI-2). As a comparison, we also included baseline conditions where we created AI-Agents using simple adjectives. To validate these AI-agents, we prompted AI-agents to complete a criterion measure (i.e., the Mini-Marker test) and compared their responses to human participants' responses with matching personalities. In other words, we assess whether AI-Agents’ responses align with human participants’ responses across different AI-Agent conditions. We hypothesize that AI agents created using the BFI-2 will capture more nuances in personalities and thus show greater alignment with human responses than those created using simple adjectives.

In Study 3, we further validate the AI-Agents by prompting them to respond to real-life moral and risk-taking vignettes and examining the extent to which their responses align with human participants’ responses. We again hypothesize that AI agents created using the BFI-2 will align better with human responses than those created using simple adjectives.

Collectively, these three studies form a comprehensive investigation into the creation, validation, and application of AI-Agents with psychometrically sound personality traits. We hope to provide researchers with a powerful new tool for conducting personality and social behavior research at scale, while maintaining high standards of validity and reliability.

\subsection{Data Availability Statement}
All data, research materials, and analysis code have been made publicly available through GitHub (\url{https://github.com/muhua-h/Psychometrics4AI}). Study 3 was preregistered on the Open Science Framework prior to data collection (\url{https://osf.io/4t9bf}) and received ethical approval from the Ethics Review Board at [Institution Name] (ID: e2024-221). The sample size, exclusion criteria, and analysis plan for Study 3 were specified in the preregistration.

\section{Study 1: Semantic Representation of Personality Constructs in LLM}
Study 1 aimed to establish a foundational understanding of the semantic nuances inherent in personality tests and constructs using embedding. Because the behavior of AI-Agents is ultimately driven by embeddings, this initial analysis serves to validate the approach and set the stage for subsequent studies.

\subsection{Methods}
Our analysis incorporated content from widely recognized Big Five tests, such as the BFI \parencite{john_big_1991}, BFI-2 \parencite{soto_next_2017}, Mini-Markers \parencite{saucier_mini-markers_1994}, BFAS \parencite{deyoung_between_2007} and NEO-FFI \parencite{costa_neo_1989}. We extracted domain-specific content (i.e., test items) from these tests and processed it through OpenAI's advanced text-embedding model (\texttt{text-embedding-3-large}). Text embeddings are commonly used in natural language processing; they allow researchers to represent text data  (such as words, phrases, sentences, or even entire documents) in a numeric format (i.e., vector) that quantifies the semantic relationships between words, phrases, or entire documents. In the context of our study, each personality test item is converted into a high-dimensional vector (3072 dimensions for the model we used) that captures its semantic meaning.

 To illustrate this concept, consider three items from different Big Five tests:
\begin{itemize}
    \item[(a)] ``Is outgoing, sociable'' (BFI-2, Extraversion)
    \item[(b)] ``Extraverted'' (Mini-Markers, Extraversion)
    \item[(c)] ``Is original, comes up with new ideas'' (BFI-2, Openness)
\end{itemize}
Items (a) and (b), despite their different wording, are semantically similar and would be represented by vectors close to each other in the high-dimensional space, with a similarity of $.65$ as measured by \textit{cosine distance}. By contrast, item (c) measures a different construct, although it follows the same wording style as item (a). Item (c)'s similarity to item (a) is $.33$, and to item (b) $.23$. This example shows the extent to which embeddings capture convergent and divergent psychological constructs. 


\subsection{Embeddings Techniques}
To analyze these embeddings, we employed two techniques, namely, cosine similarity and \textit{t}-Distributed Stochastic Neighbor Embedding (\textit{t}-SNE).

\paragraph{Cosine similarity} This measure quantifies the semantic similarity between two embedded texts, analogous to how correlation coefficients measure the relationship between variables in psychological research. Just as a correlation of 1 indicates a perfect positive relationship, a cosine similarity of 1 indicates identical semantic meaning. In contrast, a cosine similarity of 0 suggests no semantic relationship, similar to a correlation of 0 indicating no linear relationship between variables.

In the context of personality assessment, cosine similarity can be thought of as a measure of construct overlap between items or scales. For instance, high cosine similarity between items from different tests (e.g., ``Is outgoing, socioable'' from BFI2 and ``Extraverted" from Mini-Markers) would suggest they are tapping into the same underlying construct, similar to how high inter-item correlations within a scale indicate internal consistency in classical test theory. 

\paragraph{\textit{t}-SNE} 
We applied \textit{t}-SNE \parencite{van2008visualizing} to item embeddings to create a two-dimensional visualization of their similarity structure. t-SNE is a non-linear dimensionality reduction method that maps high-dimensional data to a low-dimensional space while preserving local relationships between data points. In our context, the t-SNE provides an intuitive visual map of semantic neighbourhoods among personality test items, which provides a good complement to the cosine similarity analysis. Items from the same Big Five domain tend to cluster together in the plot, and cross-domain proximities become easier to spot visually. 
In our \textit{t}-SNE visualizations, items that cluster together can be interpreted as measuring similar constructs\footnote{Because t-SNE is intended for visualization rather than formal measurement modelling, the plots are used here descriptively to illustrate patterns in the embedding space; the axes have no intrinsic psychological meaning.}.

The use of embedding in this study offers several advantages for personality research. By capturing semantic relationships between test items, we can potentially uncover subtle distinctions between various personality assessments that might not be apparent through traditional psychometric methods. This approach provides a data-driven complement to classical item analysis techniques, potentially informing more refined personality models and enhancing our understanding of the semantic overlap between different personality measures.

\begin{figure}[htbp]
\caption{Cosine Similarity Between Personality Tests: Overall Average and Domain-Specific Comparisons}
\centering
\includegraphics[width=1\textwidth]{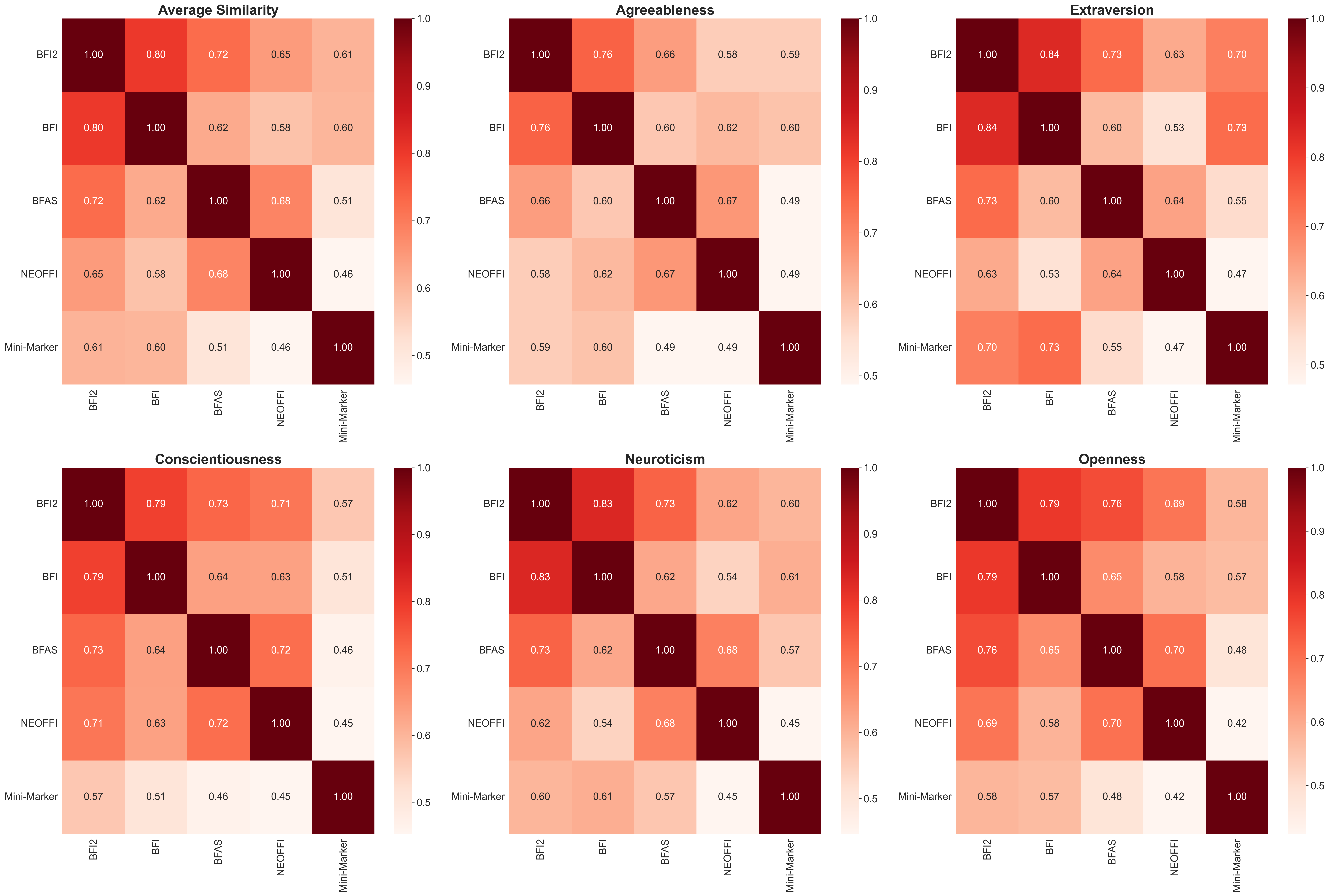}
\label{fig:cosine_similarity}
\end{figure}

\subsection{Results and Discussion}
Figure \ref{fig:cosine_similarity} illustrates the cosine similarity between different personality tests across the Big Five domains. The top-left subplot displays the average similarity across all domains, indicating that most tests generally exhibit moderate to high cosine similarity (above 0.51), except for Mini-Markers and NEO-FFI. The Mini-Markers test consistently shows a relatively lower cosine similarity with other scales across all domains, which may be attributed to its unique design approach: while other tests employ full statements, phrases, or questions for each item (e.g., ``Is complex, a deep thinker''), the Mini-Markers test exclusively uses adjectives (e.g., philosophical''). The remaining subplots, each corresponding to a specific Big Five domain, reveal similar patterns of cosine similarity between tests. This consistency across domains suggests that semantic relationships between different personality tests are relatively stable, regardless of the specific trait being measured. However, varying degrees of similarity observed between different test pairs highlight the nuanced differences in how each instrument operationalizes and measures personality constructs within the Big Five framework.

\begin{figure}[ht]
\caption{Two-Dimensional Projection of Big Five Personality Test Domain Embeddings Using t-SNE}
\centering
\includegraphics[width=.8\textwidth]{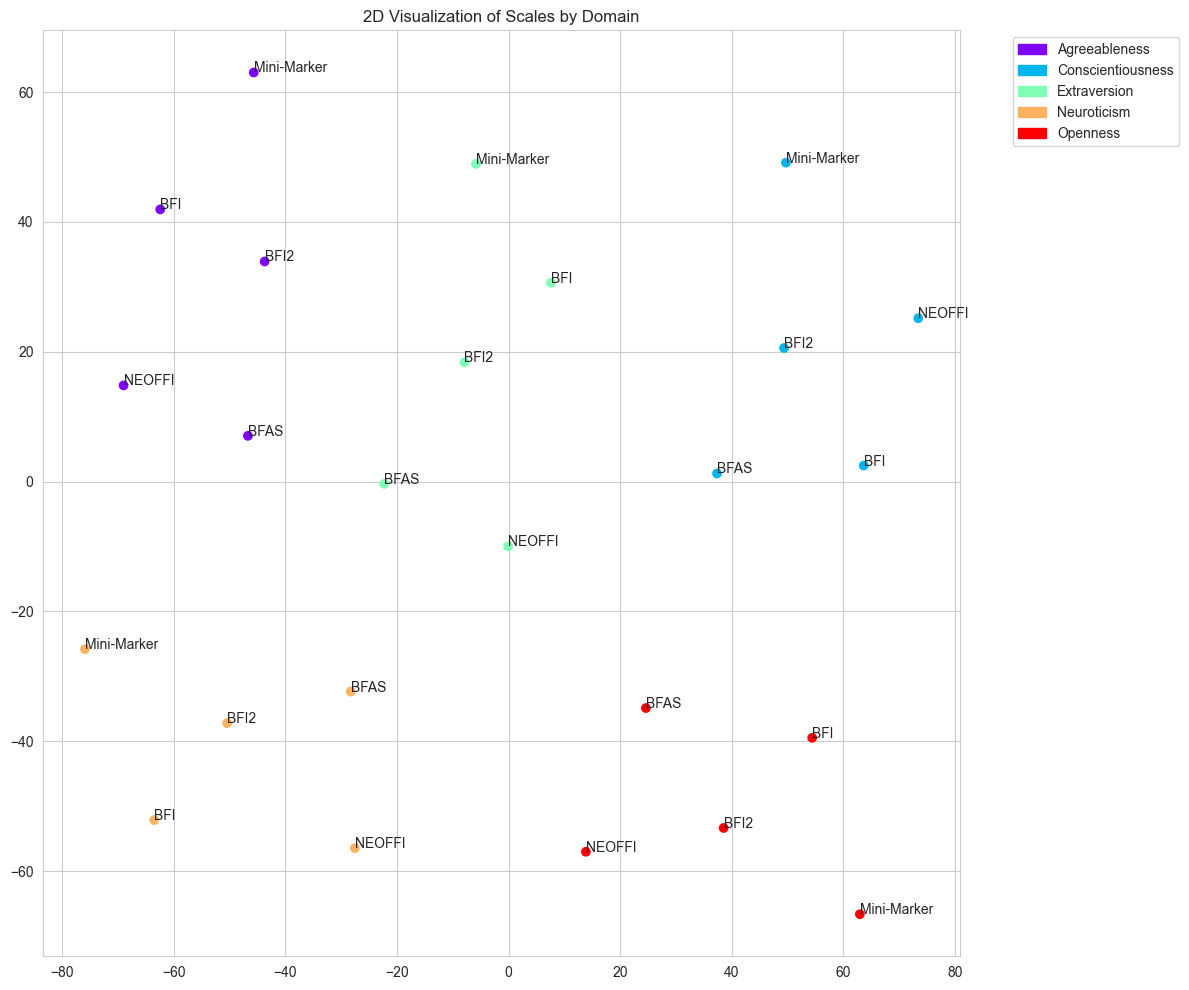}
\label{fig:tsne_projection}
\end{figure}


Figure \ref{fig:tsne_projection} shows a two-dimensional projection of the tests' domain embeddings using \textit{t}-SNE\footnote{Although \textit{t}-SNE is typically applied to larger datasets, it can still be used with smaller datasets for illustrative purposes. Here, it is employed solely for visualization, not for statistical inference. The axes for \textit{t}-SNE plot are usually not labeled, because the units of the axes are not directly related to any measurement from the original data. Instead, the axes represent a space constructed to best represent similarities and differences between data points.}, which complements the statistical results by offering an intuitive view of how items from different tests are positioned relative to each other in the embedding space.  Each point represents a personality test item, and items from the same Big Five domain tend to cluster together, mirroring within-domain similarities shown in Figure \ref{fig:cosine_similarity}. This visualization reveals distinct clustering patterns for each Big Five personality domain, with the spatial arrangement of these clusters providing insight into the semantic relationships between personality constructs. For instance, items assessing Agreeableness are concentrated in the upper left quadrant, while those measuring Openness are predominantly situated in the lower right quadrant. The remaining domains (i.e., Extraversion, Conscientiousness, and Neuroticism) form separate, distinguishable clusters within the projection space.

These results suggest that different personality tests tap into highly similar and consistent constructs, despite variations in item wording and test structure. This finding serves as the foundation for subsequent studies, which further examine Agents' understanding of the underlying semantic associations during the personality assignment, adaptation, and reflection process. 

\section{Study 2: Creating and Validating AI-Agents with Personalities}

Considering the findings of Study 1, which indicated a relatively low semantic similarity between the BFI-2 and the Mini-Markers test, in Study 2, we aim to create AI-Agents using the BFI-2 and then validate the personality assignments using the Mini-Markers test. The rationale is that if this method supports two fairly distinct psychometric tests, it should generalize to other tests with greater semantic similarity.

Study 2 consists of two parts: Study 2a and Study 2b. In Study 2a, we discuss how to use prompt engineering to create AI-agents with personalities and examine the alignment between AI-agents' responses on the Mini-Markers test responses and those of human participants collected in a previous study. In Study 2b, we demonstrate how to create and validate AI-Agents representative of human participants' sample data using only summary statistics from a previous study, offering an efficient way to create AI-agents without collecting new human data.

\subsection{Study 2a}

\subsubsection{Method}

\paragraph{Prompt Enginneering}

We included four prompt conditions for creating AI-Agents with personalities (see Table \ref{tab:examplepromtps}). Two conditions involved using the BFI-2  in either Likert or Expanded format, which we refer to as \textit{BFI-2-Likert} and \textit{BFI-2-Expanded}, respectively. The BFI-2-Likert is the original BFI-2 by \textcite{soto_next_2017}, in which participants indicated their agreement with statements measuring the Big Five traits. The BFI-2-Likert, however, is susceptible to response bias and careless responding; therefore, \textcite{Zhang_Huang_2025} developed the BFI-2-Expanded version, replacing the agree-disagree response options with complete sentences. For example, for the Expanded format,  the ``strongly agree'' option for the item ``Is outgoing, sociable'' is replaced with ``I am very outgoing, sociable''. 
For prompt engineering, the Expanded format produces prompts that sound more natural and straightforward. As shown in Table \ref{tab:examplepromtps}, the BFI-2-Likert prompt requires an additional instruction for interpreting numeric numbers indicating agreement levels, whereas the BFI-2-Expanded prompt simply describes the trait levels in complete sentences.

We also included two baseline prompt conditions (Simple-Binary and Elaborated-Binary), designed to mimic the approach of previous studies that used simple adjectives to create AI-Agent personas \textcite{jiang_personallm_2024, park_generative_2023, serapio2023personality}. In the Simple-Binary condition, prompts were binary Big Five statements indicating whether the AI-Agent was high or low on each trait. The Elaborated-Binary condition added descriptors such as   ``outgoing'' and ``compassionate'' to further elaborate on the traits. Compared with these baseline conditions, we aim to show that the BFI-2-Likert or BFI-2-Expanded condition create AI-Agents with more nuanced personalities and aligns better with human responses.

\begin{table}[htbp]
   \caption{Example Prompts for Creating AI-Agents With Personalities}
    \label{tab:examplepromtps}
\scriptsize
    \setlength{\tabcolsep}{4pt}
      \renewcommand{\arraystretch}{1.3} 
\begin{tabular}{>{\raggedright\arraybackslash}p{8cm}>{\raggedright\arraybackslash}p{8cm}} \hline 
\textbf{Simple-Binary} &\textbf{Elaborated-Binary}\\ \hline
 \\
\#\#\# Context\#\#\# 
You are participating in a personality psychology study. You have been assigned personality traits. 

\bigskip 
\#\#\#  Your Personality \#\#\#  \newline 
You are high in Extraversion. You are high in Agreeableness. You are high in Conscientiousness. You are low in Neuroticism. You are high in Openness.

&   \#\#\# Context\#\#\# 
You are participating in a personality psychology study. You have been assigned personality traits. 

\bigskip 
\#\#\#  Your Personality \#\#\#  \newline 
You are high in Extraversion. You are outgoing, sociable, assertive, and energetic. You are high in Agreeableness. You are compassionate, cooperative, trusting, and kind to others. You are high in Conscientiousness. You are organized, responsible, hardworking, and reliable. You are low in Openness. You prefer familiar routines, practical approaches, and conventional ideas.\\ \hline 
\textbf{BFI-2-Likert}&\textbf{BFI-2-Expanded}\\\hline
\\
\#\#\# Context \#\#\# \newline 
You are participating in a personality psychology study. You have been assigned personality traits.

\bigskip

\#\#\# Instruction\#\#\# \newline 
Each number below indicates the extent to which you agree or disagree with that the statement. 1 means `Disagree Strongly', 2 means `Disagree', 3 means `Neutral', and 4 means `Agree', 5 means `Agree Strongly'.\newline 

\#\#\#  Your Personality \#\#\#  \newline 
Is outgoing, sociable: 5; Is compassionate, has a soft heart: 5; Tends to be disorganized: 2; Is relaxed, handles stress well: 3; Has few artistic interests: 2; Has an assertive personality: 4; Is respectful, treats others with respect: 5; Tends to be lazy: 4; Stays optimistic after experiencing a setback: 5; Is curious about many different things: 2; Rarely feels excited or eager: 2; Tends to find fault with others: 2; Is dependable, steady: 5; Is moody, has up and down mood swings: 4; Is inventive, finds clever ways to do things: 4; Tends to be quiet: 2; Feels little sympathy for others: 1; Is systematic, likes to keep things in order: 2; Can be tense: 3; Is fascinated by art, music, or literature: 5; Is dominant, acts as a leader: 2; Starts arguments with others: 2; Has difficulty getting started on tasks: 4; Feels secure, comfortable with self: 5; Avoids intellectual, philosophical discussions: 2; Is less active than other people: 2; Has a forgiving nature: 5; Can be somewhat careless: 4; Is emotionally stable, not easily upset: 4; Has little creativity: 4; Is sometimes shy, introverted: 4; Is helpful and unselfish with others: 5; Keeps things neat and tidy: 3; Worries a lot: 4; Values art and beauty: 4; Finds it hard to influence people: 2; Is sometimes rude to others: 3; Is efficient, gets things done: 4; Often feels sad: 1; Is complex, a deep thinker: 2; Is full of energy: 5; Is suspicious of others' intentions: 2; Is reliable, can always be counted on: 4; Keeps their emotions under control: 4; Has difficulty imagining things: 2; Is talkative: 5; Can be cold and uncaring: 1; Leaves a mess, doesn't clean up: 2; Rarely feels anxious or afraid: 2; Thinks poetry and plays are boring: 2; Prefers to have others take charge: 1; Is polite, courteous toward others: 5; Is persistent, works until the task is finished: 4; Tends to feel depressed, blue: 3; Has little interest in abstract ideas: 4; Shows a lot of enthusiasm: 5; Assumes the best about people: 5; Sometimes behaves irresponsibly: 4; Is temperamental, gets emotional easily: 1; Is original, comes up with new ideas: 1.

    &  
    \#\#\#  Context \#\#\# \newline 
You are participating in a personality psychology study. You have been assigned personality traits. 
\bigskip

\#\#\#  Your Personality \#\#\#  

I am very outgoing, sociable. I am very compassionate, almost always soft-hearted. I am fairly organized. I am somewhat relaxed, handle stress somewhat well. I have some artistic interests. I am quite assertive. I am very respectful almost always treat others with respect. I am often lazy. I stay very optimistic after experiencing a setback. I am curious about few things. I often feel excited or eager. I rarely find fault with others. I am very dependable, steady. I am fairly moody often have up and down mood swings. I am fairly inventive, often find clever ways to do things. I am rarely quiet. I feel a great deal of sympathy for others. I am not particularly systematic rarely keep things in order. I am sometimes tense. I am very much fascinated by art, music or literature. I am fairly submissive, often act as a follower. I rarely start arguments with others. I have a fair amount of difficulty getting started on tasks. I feel very secure, comfortable with self. I typically seek out intellectual, philosophical discussions. I am somewhat more active than other people. I have a very forgiving nature. I am often careless. I am fairly emotionally stable quite hard to upset. I have little creativity. I am often shy, introverted. I am very helpful and unselfish with others. I sometimes keep things neat and tidy. I worry quite a lot. I value art and beauty quite a bit. I find it fairly easy to influence people. I am sometimes rude to others. I am fairly efficient, get things done fairly quickly. I almost never feel sad. I am not particularly complex rarely a deep thinker. I am almost always full of energy. I am quite trusting of others’ intentions. I am fairly reliable, can usually be counted on. I usually keep my emotions under control. I have a bit of difficulty imagining things. I am very talkative. I am very warm and caring. I rarely leave a mess usually clean up. I often feel anxious or afraid. I think poetry and plays are fairly interesting. I strongly prefer to take charge. I am very polite and courteous to others. I am fairly persistent, usually work until the task is finished. I sometimes feel depressed and blue. I have little interest in abstract ideas. I show a lot of enthusiasm. I almost always assume the best about people. I often behave irresponsibly. I am not at all temperamental almost never get emotional. I am not at all original almost never come up with new ideas.\\ \hline 
\end{tabular}
\end{table}

\paragraph{Data}
To examine alignment between AI-Agents responses with those from human participants, we repurposed data collected by \textcite{soto_next_2017} wherein participants (\textit{N} = 438) responded to multiple Big Five tests, including: (1) BFI-2: a sophisticated 60-item Likert test designed to capture the comprehensive hierarchical structure of personality, where each domain has three facets and (2) Mini-Markers: a straightforward, 40-item test consisting of phenotypic Big Five descriptive adjectives \parencite{saucier_mini-markers_1994}.  Participants’ BFI2 responses will serve as the training (input) data, while their Mini-Markers responses (output) will be used for validation. The human data \parencite{soto_next_2017}  will serve as a reference for our comparison.   

\paragraph{Procedure}

Using the four types of prompts in Table \ref{tab:examplepromtps}, we created AI-Agents across five popular LLMs, including 1) GPT-3.5-Turbo (01-25), 2) GPT-4-Turbo (04-09), 3) GPT-4o (2024-11-20), 4)  DeepSeek-V3 (2024-12-26), and 5) Llama-3.3-70B-Instruct (2024-12-06). For each LLM and each prompt condition, we initiated 438 AI-Agents, each corresponding to a human participant's data collected from \textcite{soto_next_2017}. More specifically, for the BFI-2-Likert and BFI-2-Expanded conditions, AI-Agents were created by inputting prompts that matched the BFI-2 scores of the corresponding human participants. For the Baseline-Simple and Baseline-Elaborated conditions, participants' trait scores were dichotomized as high (average score $> 2.5$) or low (average score $< 2.5$) for each of the Big Five traits, and AI-Agents were then created using these dichotomized scores. Across all conditions, the LLM temperature was set to the default value of 1.0 to balance coherence and diversity, and other sampling parameters (e.g., top-p, frequency, and presence penalties) were kept at their defaults. 


After creating AI-Agents for each LLM and prompt condition, we prompted them to complete the Mini-Markers test, which are 40 adjectives describing the Big Five traits  \parencite{saucier_mini-markers_1994}. The full prompt for this task is provided in the Appendix A. In this prompt, AI-Agents were instructed to consider the Mini-Marker items carefully based on their assigned personality (see Appendix).

\subsubsection{Data Analysis}

\paragraph{Comparing Mini-Marker Scores Between AI-Agents and Humans}
For each LLM, prompt condition, and personality trait, we conducted correlational analyses, paired-sample $t$-tests, and Kolmogorov-Smirnov (KS) tests to compare the Mini-Markers responses of AI-agents with those of humans who had matching underlying BFI-2 personalities. Correlational analyses examined the linear relationships between AI-agent and human responses, paired-sample $t$-tests compared their mean Mini-Markers scores, and KS tests compared their score distributions. For all tests, Bonferroni corrections are used to control for family-wise Type I error rate. 

\paragraph{Correlational Analyses Between BFI-2 and Mini-Markers}
In addition, for each LLM and personality trait, we computed the correlation between the Big Five and the Mini-Markers. Specifically, for the Likert and Expanded prompt conditions, we correlated the Mini-Marker output scores with the BFI-2-Likert and -Expanded input scores (via prompts), respectively.  

For the Baseline-Simple and Baseline-Elaborate conditions, we computed two types of correlations. 
First, we correlate the original BFI-2-Likert scores (from which the binary inputs were derived) with the Mini-Markers output scores. Second, we correlate the binary Big Five input scores with the Mini-Markers scores. We add the second set of correlations because LLMs only received the binary inputs in those two conditions without access to the original BFI-2-Likert scores. 
By comparing these correlations from AI-Agents to those from humans (i.e., human participants' correlation between BFI-2 and Mini-markers), we can determine whether AI-agents can help researchers recover correlations between variables in research. 

\paragraph{Factor Analyses}
Finally, we conducted Confirmatory Factor Analyses (CFA)
\footnote{CFA is a widely used psychometric method for testing whether a set of questionnaire items reflects an expected structure of psychological traits\parencite{kline2014easy}. For example, in the Big Five model, items about being “outgoing” or “sociable” should cluster together as indicators of Extraversion. CFA formally tests such hypotheses by estimating how strongly each item loads onto a latent factor (e.g., Extraversion) and by providing indices of overall model fit (e.g., CFI, RMSEA, SRMR). Good fit means the data are consistent with the hypothesized trait structure, while poor fit suggests the responses may not capture the intended personality dimensions. In our study, CFA allows us to evaluate whether AI-Agent responses reproduce the same underlying Big Five structure observed in human data.}using the \textit{lavaan} package in R \parencite{rosseel2012lavaan}. For each LLM, prompt condition, and personality trait, we fitted a one-factor model and compute the one-factor reliability coefficient (a.k.a., Omega).  To evaluate the fit of the two CFA models, we used the chi-square test of fit and three approximate fit indices with common cutoff points: 1) the comparative fit index (CFI), with a value above .90 indicating a reasonable fit and above .95 indicating a very good fit; 2) the root mean square error of approximation (RMSEA), with a value of less than .08 indicating reasonable fit and less than .05 indicating very good fit; and 3) the standardized root mean square residual (SRMR),  with a value of less than .08 indicating reasonable fit and less than .05 indicating very good fit. 

\subsubsection{Results and Discussion}

\paragraph{Comparing Mini-Marker Scores Between AI-Agents and Humans}

\begin{figure}[htbp]
    \caption{Mini-Markers' Conscientiousness Scores For Human and AI-Agents}
    \label{fig:dist}
    \includegraphics[width=1.1\textwidth]{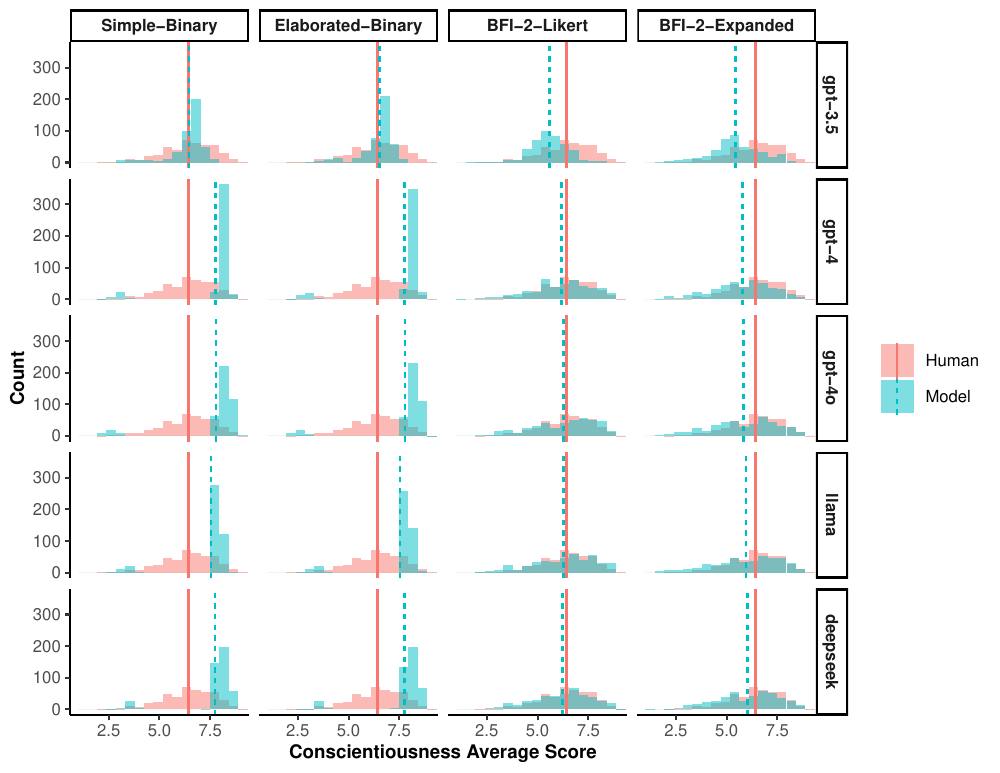}

\footnotesize
\emph{Note.} The red line indicates the mean of the human responses. The dotted blue lines indicate the means of AI-Agents across conditions.
\end{figure}

Table \ref{tab:pairedttest} shows the mean differences in average Mini-Marker scores between AI-Agents and human responses across prompt and LLM conditions. Table \ref{tab:ks} presents the KS test statistics, which measure the maximum absolute differences between the distributions of AI-Agents and human responses. Figure \ref{fig:dist} shows graphs visualizing the mean and distributional differences for the Conscientiousness average scores (see Supplementary Materials for the graphs for the other four traits).

As shown in Tables \ref{tab:pairedttest} and \ref{tab:ks}, the means and distributions differed significantly between AI-Agents and human responses in the Simple- and Elaborated-Binary prompt conditions. As illustrated in Figure \ref{fig:dist}, except for GPT-3.5, AI-Agents' Mini-Markers scores in these conditions were essentially binary, with most responses clustering at the very high end and some at the very low end, resulting in higher mean scores than those observed in humans. In other words, for GPT-4, GPT-4o, Llama, and DeepSeek, when the assigned personality prompts consisted of simple binary adjectives, the models interpreted the underlying assigned personality as binary, either very high or very low on a given trait. By contrast, for humans, regardless of whether traits were measured dichotomously or continuously, underlying distributions were continuous and typically normally distributed. Interestingly, GPT-3.5, the older and smaller model, tended to generate more moderate and variable results even under the Simple-Binary and Elaborated-Binary conditions. This suggests that as the LLMs grow bigger and their capability improves, the models become more sensitive to personality steering prompts, making simple personality descriptions less suitable to simulate human populations.  

For the BFI-2-Likert and -Expanded prompt conditions, although most LLMs still produced AI-Agents with mean scores significantly different from those of humans (see Table \ref{tab:pairedttest}), many generated distributions that closely resembled those of humans. As shown in Table \ref{tab:ks} and Figure \ref{fig:dist}, the score distributions of GPT-4, GPT-4o, Llama, and DeepSeek were particularly similar to human distributions, especially under the BFI-2-Likert condition. Interestingly, GPT-3.5 once again diverged from this pattern: consistent with its behavior in the Simple- and Elaborated-Binary conditions, its AI-Agent responses were more centered around the mid-point (i.e., around 5) and exhibited a more bell-shaped distribution.

Taken together, these results indicate that the bigger and newer LLMs (GPT-4, GPT-4o, Llama, and DeepSeek) tend to reproduce the response outputs more consistent with the prompt inputs. Specifically, when the prompt presented the Big Five traits as either high or low (i.e., in the Simple- and Elaborated-Binary conditions), AI-Agent responses also fell at the extremes; when the prompt presented varying levels of the traits, responses formed a continuum. Consequently, newer LLMs showed poor alignment with human responses under the Simple- and Elaborated-Binary conditions but much better alignment under the BFI-2-Likert and -Expanded conditions. By contrast, the older GPT-3.5 consistently generated more moderate responses across prompt conditions, resulting in mediocre alignment across conditions.

\begin{table}[htbp]
\caption{Mean Differences between AI-Agents and Human Responses on the Mini-Markers}
\label{tab:pairedttest}
\small
\begin{tabular}{llccccc}
\toprule
\textbf{Condition} & \textbf{LLM} & \textbf{O} & \textbf{C} & \textbf{E} & \textbf{A} & \textbf{N} \\
\midrule
Simple-Binary & GPT-3.5   & -0.58 & \cellcolor{gray!20}{-0.04} & -0.79 & -0.48 & 0.94 \\
Simple-Binary & GPT-4     & -1.42 & -1.34 & -1.78 & -1.55 & \cellcolor{gray!20}{0.03} \\
Simple-Binary & GPT-4o    & -1.88 & -1.38 & -1.94 & -1.55 & -0.93 \\
Simple-Binary & Llama     & -1.19 & -1.12 & -1.91 & -1.55 & 0.77 \\
Simple-Binary & DeepSeek  & -1.54 & -1.32 & -1.84 & -1.82 & 0.35 \\
Elaborated-Binary & GPT-3.5   & -0.59 & \cellcolor{gray!20}{-0.11} & -0.79 & -0.52 & 0.95 \\
Elaborated-Binary & GPT-4     & -1.42 & -1.34 & -1.78 & -1.55 & \cellcolor{gray!20}{-0.08} \\
Elaborated-Binary & GPT-4o    & -1.88 & -1.36 & -1.94 & -1.55 & -0.91 \\
Elaborated-Binary & Llama     & -1.22 & -1.13 & -1.91 & -1.57 & 0.77 \\
Elaborated-Binary & DeepSeek  & -1.58 & -1.35 & -1.81 & -1.77 & 0.26 \\
BFI-2-Likert & GPT-3.5   & 1.71 & 0.84 & 0.43 & 0.86 & 0.95 \\
BFI-2-Likert & GPT-4     & 1.22 & 0.24 & 0.47 & 0.29 & 0.26 \\
BFI-2-Likert & GPT-4o    & 0.20 & 0.15 & 0.35 & 0.15 & 0.20 \\
BFI-2-Likert & Llama     & \cellcolor{gray!20}{0.04} & 0.14 & \cellcolor{gray!20}{-0.01} & \cellcolor{gray!20}{-0.12} & 0.33 \\
BFI-2-Likert & DeepSeek  & 0.40 & 0.19 & \cellcolor{gray!20}{0.02} & 0.30 & \cellcolor{gray!20}{0.07} \\
BFI-2-Expanded & GPT-3.5   & 0.44 & 0.96 & 0.29 & 0.29 & 0.43 \\
BFI-2-Expanded & GPT-4     & 0.17 & 0.65 & 0.22 & 0.22 & 0.44 \\
BFI-2-Expanded & GPT-4o    & \cellcolor{gray!20}{0.02} & 0.57 & 0.29 & 0.53 & -0.34 \\
BFI-2-Expanded & Llama     & -0.13 & 0.46 & \cellcolor{gray!20}{-0.02} & 0.25 & 0.29 \\
BFI-2-Expanded & DeepSeek  & -0.30 & 0.39 & \cellcolor{gray!20}{-0.14} & 0.20 & 0.35 \\
\bottomrule
\end{tabular}
\footnotesize

\medskip
\emph{Note}: Unshaded cells indicate significant paired-sample $t$-tests comparing 

AI-Agent and human responses, whereas gray-shaded cells indicate

non-significant results. Significance levels were adjusted using 

Bonferroni corrections to control the family-wise Type I error rate. 
\end{table}
\normalsize

\begin{table}[htbp]
\caption{Kolmogorov-Smirnov Test Statistic Comparing AI-Agents and Human Score Distributions for the Mini-Markers}
\label{tab:ks}\small
\begin{tabular}{llccccc}
\toprule
\textbf{Condition} & \textbf{LLM} & \textbf{O} & \textbf{C} & \textbf{E} & \textbf{A} & \textbf{N} \\
\midrule
Simple-Binary & GPT-3.5 & 0.47 & 0.23 & 0.56 & 0.42 & 0.45 \\
Simple-Binary & GPT-4   & 0.77 & 0.78 & 0.74 & 0.84 & 0.25 \\
Simple-Binary & GPT-4o  & 0.87 & 0.76 & 0.76 & 0.79 & 0.59 \\
Simple-Binary & Llama   & 0.77 & 0.72 & 0.78 & 0.85 & 0.32 \\
Simple-Binary & DeepSeek& 0.79 & 0.76 & 0.75 & 0.87 & 0.27 \\
Elaborated-Binary & GPT-3.5 & 0.44 & 0.21 & 0.55 & 0.42 & 0.46 \\
Elaborated-Binary & GPT-4   & 0.77 & 0.78 & 0.75 & 0.84 & 0.28 \\
Elaborated-Binary & GPT-4o  & 0.87 & 0.77 & 0.76 & 0.80 & 0.59 \\
Elaborated-Binary & Llama   & 0.78 & 0.71 & 0.78 & 0.85 & 0.33 \\
Elaborated-Binary & DeepSeek& 0.79 & 0.75 & 0.75 & 0.85 & 0.27 \\
BFI-2-Likert & GPT-3.5   & 0.62 & 0.39 & 0.22 & 0.44 & 0.37 \\
BFI-2-Likert & GPT-4     & 0.42 & \cellcolor{gray!20}{0.11} & 0.19 & 0.14 & \cellcolor{gray!20}{0.07} \\
BFI-2-Likert & GPT-4o    & 0.13 & \cellcolor{gray!20}{0.10} & 0.20 & \cellcolor{gray!20}{0.12} & \cellcolor{gray!20}{0.09} \\
BFI-2-Likert & Llama     & \cellcolor{gray!20}{0.07} & \cellcolor{gray!20}{0.09} & 0.13 & 0.21 & \cellcolor{gray!20}{0.11} \\
BFI-2-Likert & DeepSeek  & 0.19 & \cellcolor{gray!20}{0.08} & \cellcolor{gray!20}{0.09} & 0.15 & \cellcolor{gray!20}{0.06} \\
BFI-2-Expanded & GPT-3.5 & 0.13 & 0.38 & \cellcolor{gray!20}{0.12} & 0.19 & 0.21 \\
BFI-2-Expanded & GPT-4   & \cellcolor{gray!20}{0.09} & 0.21 & 0.16 & \cellcolor{gray!20}{0.12} & 0.13 \\
BFI-2-Expanded & GPT-4o  & \cellcolor{gray!20}{0.07} & 0.19 & 0.18 & 0.20 & 0.14 \\
BFI-2-Expanded & Llama   & \cellcolor{gray!20}{0.12} & 0.14 & 0.20 & 0.14 & \cellcolor{gray!20}{0.10} \\
BFI-2-Expanded & DeepSeek& 0.18 & 0.15 & 0.22 & \cellcolor{gray!20}{0.11} & 0.13 \\
\bottomrule
\end{tabular}

\footnotesize
\medskip
\emph{Note}: Unshaded cells indicate significant paired-sample $t$-tests comparing 

AI-Agent and human responses, whereas gray-shaded cells indicate

non-significant results. Significance levels were adjusted using 

Bonferroni corrections to control the family-wise Type I error rate. 
\end{table}
\normalsize

\normalsize

\paragraph{Correlational Analyses Between BFI-2 and Mini-Markers}

Table \ref{tab:bfi_convergence_all}, for the BFI-2-Likert and -Expanded conditions, shows domain-level Pearson correlations between BFI-2 input scores and Mini-Markers output scores. For the Simple- and Elaborated-Binary conditions, Table \ref{tab:bfi_convergence_all} includes two sets of correlations:  (1) correlations between the binary Big Five input scores and Mini-Markers output scores (right-hand values), and (2) correlations between the original BFI-2 Likert scores (from which binary inputs were derived) and Mini-Markers output scores (left-hand values). 
As a reference, the first row of Table \ref{tab:bfi_convergence_all} shows the correlations for human participants between their BFI-2-Likert and Mini-Markers scores (Human: O=.75, C=.84, E=.88, A=.80, N=.74; Avg=.80; $N$ = 438). 

Across all conditions, the correlation coefficients in Table \ref{tab:bfi_convergence_all} were significantly different from zero, indicating that Big Five input and Mini-Marker output were significantly related across conditions. For the BFI-2-Likert and BFI-2-Expanded conditions, across LLMs and traits, correlations between the BFI-2 input and the Mini-Makers output scores were generally high  (Likert Avgs: GPT-3.5=.75 GPT-4=.86, GPT-4o=.91, Llama=.90, DeepSeek=.90; Expanded Avgs:  .78, .86, .90, .90, .89). 

On the other hand, for Simple- and Elaborated-Binary conditions, across LLMs and traits, right-hand correlations (between binary Big Five input and Mini-Markers output scores) were generally high, but left-hand correlations (between the original BFI-2 and Mini-Markers scores) were substantially lower. For example, under the Simple-Binary for GPT-4, Conscientiousness' left-hand and right-hand correlations were .58 and .94, respectively,  Agreeableness were .35 and .91; for GPT-4o, Openness were .45 and .81 and Agreeableness were .42 and .88. This split mirrors the earlier item-score comparison: LLMs tend to produce quasi-binary Mini-Marker outputs under binary prompts, which (1) mismatch continuous human Likert scores, which depresses left-hand correlations, but (2) aligns tightly with the binary inputs, elevating right-hand correlations. The high right-hand values indicate successful tracking of the prompted high/low signal, whereas the low left-hand values remind us that such tracking does \emph{not} necessarily reproduce the underlying continuous human trait covariation. The inflated correlations are a warning for researchers who attempt to approximate human population distribution using binary personality assignment. 

Across models, newer and more capable models achieved higher convergence compared to older ones. For example, average correlations for GPT-3.5, GPT-4, and GPT-4o, were .75, 86, .91 for BFI-2-Likert condition, and .78, .86, .90 for BFI-2-Expanded condition. By domain, Conscientiousness and Extraversion are consistently strongest for newer models in richer formats (e.g., Likert C$\approx$.91-.93; E$\approx$.92-.94), Agreeableness was the main laggard for GPT-3.5 (Likert A=.67; Expanded A=.64), and Openness showed the widest model spread under Likert (GPT-3.5=.60 vs Llama=.89), with Expanded partially narrowing gaps for older models. 

Regarding human-AI alignment, Table \ref{tab:bfi_convergence_all} shades in gray the cells that are not statistically different from the human reference (adjusted with Bonferroni-corrected Fisher-z); unshaded cells indicate significant differences. There are few noticeable patterns of results. First, for the BFI-2-Likert and -Expanded conditions, the AI-Agents and humans had very good alignment for the Openness and Extraversion traits and some alignment for the Conscientiousness and Agreeableness traits, with the Expanded format showing better alignment than the Likert format.
Second, for Simple- and Elaborate-Binary conditions,  left-handed correlations (between the original BFI-2 and Mini-Markers scores), except for the Neuroticism trait, showed very poor alignment, especially for newer LLMs. On the other hand, right-hand correlations (between the binary Big Five input and Mini-Markers output scores) showed good alignment for the Openness and Extraversion traits but were inflated relative to human correlations for the other traits. These results again underscore that newer LLMs reliably track low/high signals embedded in the prompt inputs, but in doing so they fail to reproduce the subtler, continuous covariation that characterizes human personality traits.

\begin{table}[htbp] 

\setlength{\tabcolsep}{4pt}        
\renewcommand{\arraystretch}{1.1}  

\begin{threeparttable}
\caption{Correlation between BFI-2 Input and the Mini-Markers Output (Study 2a)}
\label{tab:bfi_convergence_all}

{\singlespace
\footnotesize 
\begin{tabular}{llcccccc}
\toprule
Condition & Model & O & C & E & A & N & Avg \\
\midrule
Human &  & .75 & .84 & .88 & .80 & .74 & .80 \\ \hline
Simple-Binary & GPT-3.5 & .37/\shadesecond{.70} & .55/\shadesecond{.79} & .67/\shadesecond{.88} & .29/\shadesecond{.74} & .56/.57 & .49/\shadesecond{.73} \\
Simple-Binary & GPT-4 & .42/\shadesecond{.79} & .58/.94 & .67/\shadesecond{.89} & .35/.91 & \shadefirst{.76}/.88 & .56/.88 \\
Simple-Binary & GPT-4o & .45/\shadesecond{.81} & .62/.91 & .69/\shadesecond{.88} & .42/.88 & \shadefirst{.77}/.90 & .59/.88 \\
Simple-Binary & Llama & .42/\shadesecond{.82} & .58/.92 & .67/\shadesecond{.88} & .33/.88 & \shadefirst{.78}/.87 & .56/.87 \\
Simple-Binary & Deepseek & .36/\shadesecond{.76} & .58/.92 & .67/\shadesecond{.88} & .37/.89 & \shadefirst{.78}/.88 & .55/.87 \\
Elaborated-Binary & GPT-3.5 & .37/\shadesecond{.71} & .58/\shadesecond{.77} & .67/\shadesecond{.87} & .28/\shadesecond{.73} & .54/.58 & .49/\shadesecond{.73} \\
Elaborated-Binary & GPT-4 & .42/\shadesecond{.77} & .58/.93 & .68/\shadesecond{.89} & .37/.92 & \shadefirst{.77}/.88 & .56/.88 \\
Elaborated-Binary & GPT-4o & .44/\shadesecond{.82} & .61/.92 & .69/\shadesecond{.88} & .39/.88 & \shadefirst{.77}/.90 & .58/.88 \\
Elaborated-Binary & Llama & .41/\shadesecond{.79} & .57/.92 & .67/\shadesecond{.89} & .36/.86 & \shadefirst{.78}/.87 & .56/.87 \\
Elaborated-Binary & Deepseek & .38/\shadesecond{.75} & .57/.92 & .68/\shadesecond{.88} & .37/.88 & \shadefirst{.78}/.88 & .55/\shadesecond{.86} \\
BFI-2-Likert & GPT-3.5 & .60 & \cellcolor{gray!20}{.85} & \cellcolor{gray!20}{.86} & .67 & \cellcolor{gray!20}{.79} & \cellcolor{gray!20}{.75} \\
BFI-2-Likert & GPT-4 & \cellcolor{gray!20}{.69} & .91 & .92 & .90 & .89 & \cellcolor{gray!20}{.86} \\
BFI-2-Likert & GPT-4o & .86 & .93 & \cellcolor{gray!20}{.92} & .90 & .92 & .91 \\
BFI-2-Likert & Llama & .89 & .91 & .94 & .87 & .92 & .90 \\
BFI-2-Likert & Deepseek & \cellcolor{gray!20}{.82} & .92 & .93 & .90 & .91 & .90 \\
BFI-2-Expanded & GPT-3.5 & \cellcolor{gray!20}{.78} & \cellcolor{gray!20}{.79} & \cellcolor{gray!20}{.86} & .64 & .83 & \cellcolor{gray!20}{.78} \\
BFI-2-Expanded & GPT-4 & \cellcolor{gray!20}{.80} & \cellcolor{gray!20}{.87} & \cellcolor{gray!20}{.90} & \cellcolor{gray!20}{.84} & .90 & \cellcolor{gray!20}{.86} \\
BFI-2-Expanded & GPT-4o & \cellcolor{gray!20}{.82} & .93 & \cellcolor{gray!20}{.92} & .90 & .91 & .90 \\
BFI-2-Expanded & Llama & .86 & .92 & \cellcolor{gray!20}{.92} & .87 & .91 & .90 \\
BFI-2-Expanded & Deepseek & \cellcolor{gray!20}{.82} & .92 & \cellcolor{gray!20}{.91} & .87 & .93 & .89 \\
\bottomrule
\end{tabular}
} 

\begin{tablenotes}[flushleft]

\footnotesize

\medskip

\textit{Note.} For the Simple- and Elaborated-Binary conditions, there are two sets of correlations:  (1) the correlations between the binary Big Five input scores and the Mini-Markers output scores (right-hand values), and (2) the correlations between the original BFI-2 Likert scores (from which the binary inputs were derived) and the Mini-Markers output scores (left-hand values). For BFI-2-Likert and -Expanded conditions, the correlations were between the BFI-2 input scores and the Mini-Markers output scores. All the correlation coefficients were signficantly different from zero at $\alpha=.05$.
Light-gray shading marks values that are \emph{not} statistically different from the Human reference after Bonferroni correction across domains ($\alpha=.05$; two-sided Fisher $z$ test on Fisher-transformed $r$); unshaded values indicate significant differences.
\end{tablenotes}

\end{threeparttable}
\end{table}

\paragraph{Factor Analyses}

Table \ref{tab:factor-analysis} shows fit measures from the factor analyses. Table \ref{tab:study2aloadings} shows selected results for factor loadings and reliability coefficients. Full results from factor analyses are provided in the Supplementary Materials. 

The most prominent pattern is that for newer LLMs (GPT-4, GPT-4o, Llama, and DeepSeek), the Simple- and Elaborated-Binary conditions yielded very good model fit and extremely high reliability, greatly exceeding those observed for the human condition (see Tables \ref{tab:factor-analysis} and \ref{tab:study2aloadings}). As shown in Table \ref{tab:study2aloadings}, with GPT-4o factor loadings and reliabilities in the binary conditions were all greater than 0.90, with many reaching 0.99. This pattern of results again reflects stronger consistency between prompt inputs and LLM outputs for the newer models; in the binary conditions, AI-agents tended to produce identical output responses across items, producing extremely high inter-item correlations. For illustration, Table \ref{tab:corMatrix} shows the correlation matrix for Neuroticism items in the Elaborated-Binary condition under GPT-4o. The correlations are uniformly high and markedly different from those based on human data, which explains the unrealistically good fit observed for these conditions.

For BFI-2-Likert and -Expanded conditions with the newer LLMs, although fit indices and reliability coefficients were lower than in binary conditions, results were overall more aligned with those from human data;  their correlation matrices also more resembled those of humans (see Tables \ref{tab:factor-analysis} and \ref{tab:study2aloadings}, \ref{tab:corMatrix}). Despite these relative similarities, notable differences remained. First, AI-Agent in these conditions showed considerably poorer model fit than human data, especially in terms of the CFI and RMSEA values, although, in some Likert conditions, SRMR values were comparable to human data (see Table \ref{tab:factor-analysis}). Second, the pattern of factor loadings differed: items that loaded especially highly for humans did not necessarily load highly for AI-Agents; nonetheless, the overall reliability coefficients were similar between AI-Agents and humans in these conditions (see Table \ref{tab:study2aloadings}).

Finally, for GPT-3.5, Simple- and Elaborated-Binary conditions did not yield as high a reliability as those for newer LLMs. The model fit were generally lower than that of human data across prompt conditions. Except for the Neuroticism subscale, reliabilities for other traits across prompt conditions were comparable to those of human data (see Table \ref{tab:study2aloadings}). The Neuroticism subscale performed poorly with GPT-3.5, yielding reliabilities as low as 0.26 (see Table \ref{tab:study2aloadings}). It seems that the reason for this poor performance is that some items such as ``relaxed'' and ``unenvious'' loaded extremely poorly under GPT-3.5's AI-Agents. Overall, GPT-3.5 performed worse than newer LLMs. 


\begin{table}[htbp] 
\caption{Selected Results for Factor Loadings and Reliability in Study 2a} \footnotesize
\label{tab:study2aloadings}
\begin{tabular}{llccccc}
\toprule 
Domain & Item & Human & \parbox{1.5cm}{\centering Simple\\Binary} & \parbox{2cm}{\centering Elaborated\\Binary} & \parbox{1.5cm}{\centering BFI-2\\Likert} & \parbox{1.7cm}{\centering BFI-2\\Expanded} \\ 
\midrule
& & &\multicolumn{4}{c}{\textbf{GPT-3.5}}\\
\multirow{9}{*}{Neuroticism} & Envious & .84 & .94 & .94 & .99 & .97 \tabularnewline
 & Fretful & .52 & .24 & .20 & .26 & .20 \tabularnewline
 & Jealous & .84 & .95 & .96 & .99& .96 \tabularnewline
 & Moody & .58 & .22 & .23 & .35 & .24 \tabularnewline
 & Relaxed & .36 & .12 & .17 & -.09 & .00 \tabularnewline
 & Temperamental & .63 & .27 & .25 & .36 & .28 \tabularnewline
 & Touchy & .52 & .28 & .35 & .57 & .36 \tabularnewline
 & Unenvious & .64 & .27 & .26 & .33 & .54 \tabularnewline
 & \textbf{Reliability} & .81 & .26 & .27 & .51 & .29 \tabularnewline
\midrule 
\multirow{9}{*}{Openness} & Complex & .37 & .65 & .70 & .79 & .65 \tabularnewline
 & Creative & .51 & .94 & .93 & .83 & .38 \tabularnewline
 & Deep & .87 & .63 & .71 & .88 & .70 \tabularnewline
 & Imaginative & .82 & .94 & .95 & .85 & .44 \tabularnewline
 & Intellectual & .51 & .88 & .91 & .72 & .96 \tabularnewline
 & Philosophical & .45 & .74 & .79 & .67 & .94 \tabularnewline
 & Uncreative & .72 & .32 & .37 & .10 & .32 \tabularnewline
 & Unintellectual & .36 & .39 & .44 & .04 & .45 \tabularnewline
 & \textbf{Reliability} & .76 & .88 & .92 & .84 & .72 \tabularnewline \hline 
& & & \multicolumn{4}{c}{\textbf{GPT-4o}}\\
\multirow{9}{*}{Neuroticism} & Envious & .84 & .98 & .98 & .69 & .94 \tabularnewline
 & Fretful & .52 & .99 & .99 & .69 & .64 \tabularnewline
 & Jealous & .84 & .98 & .98 & .71 & .94 \tabularnewline
 & Moody & .58 & .99 & .99 & .84 & .65 \tabularnewline
 & Relaxed & .36 & .98 & .98 & .61 & .61 \tabularnewline
 & Temperamental & .63 & .99 & .99 & .91 & .69 \tabularnewline
 & Touchy & .52 & .99 & .99 & .95 & .79 \tabularnewline
 & Unenvious & .64 & .98 & .98 & .68 & .94 \tabularnewline
 & \textbf{Reliability} & .81 & .99 & .99 & .89 & .82 \tabularnewline \hline 
\multirow{9}{*}{Openness} & Complex & .37 & .94 & .94 & .72 & .54 \tabularnewline
 & Creative & .51 & .97 & .98 & .60 & .35 \tabularnewline
 & Deep & .87 & .92 & .93 & .74 & .57 \tabularnewline
 & Imaginative & .82 & .98 & .99 & .62 & .34 \tabularnewline
 & Intellectual & .51 & .95 & .95 & .96 & .99 \tabularnewline
 & Philosophical & .45 & .95 & .95 & .93 & .98 \tabularnewline
 & Uncreative & .72 & .93 & .93 & .57 & .39 \tabularnewline
 & Unintellectual & .36 & .91 & .90 & .85 & .94 \tabularnewline
 & \textbf{Reliability} & .76 & .98 & .98 & .87 & .70 \tabularnewline
\bottomrule
\end{tabular}

\footnotesize
\emph{Note.} Reliability is computed based on one-factor model reliability (a.k.a., Omega coefficient). 

\end{table}

\begin{table}[htbp]
\footnotesize
\caption{Correlation Matrices Between Neuroticism Items In Selected Conditions}
\label{tab:corMatrix}
\setlength{\tabcolsep}{4pt}
\begin{tabular}{lcccccccc}
\toprule
& Envious & Fretful & Jealous & Moody & Temper- & Touchy & Relaxed & Un- \\
& &  &  &  & amental &  &  & envious  \\
\midrule
\multicolumn{9}{l}{\textbf{Human}} \\
Envious       & 1.00 &       &       &       &       &       &       &       \\
Fretful       & 0.41 & 1.00  &       &       &       &       &       &       \\
Jealous       & 0.76 & 0.40  & 1.00  &       &       &       &       &       \\
Moody         & 0.39 & 0.44  & 0.42  & 1.00  &       &       &       &       \\
Temperamental & 0.46 & 0.38  & 0.45  & 0.64  & 1.00  &       &       &       \\
Touchy        & 0.39 & 0.22  & 0.40  & 0.45  & 0.59  & 1.00  &       &       \\
Relaxed       & 0.24 & 0.42  & 0.23  & 0.33  & 0.30  & 0.20  & 1.00  &       \\
Unenvious     & 0.60 & 0.25  & 0.57  & 0.31  & 0.28  & 0.22  & 0.22  & 1.00  \\
\\
\multicolumn{9}{l}{\textbf{Elaborated-Binary with GPT-4o}} \\
Envious       & 1.00 &       &       &       &       &       &       &       \\
Fretful       & 0.98 & 1.00  &       &       &       &       &       &       \\
Jealous       & 0.99 & 0.98  & 1.00  &       &       &       &       &       \\
Moody         & 0.97 & 0.99  & 0.97  & 1.00  &       &       &       &       \\
Temperamental & 0.98 & 0.99  & 0.98  & 0.99  & 1.00  &       &       &       \\
Touchy        & 0.98 & 0.99  & 0.98  & 0.99  & 0.99  & 1.00  &       &       \\
Relaxed       & 0.96 & 0.98  & 0.96  & 0.97  & 0.98  & 0.98  & 1.00  &       \\
Unenvious     & 0.98 & 0.97  & 0.98  & 0.97  & 0.97  & 0.98  & 0.96  & 1.00  \\
\\
\multicolumn{9}{l}{\textbf{BFI-2-Likert with GPT-4o}} \\
Envious       & 1.00 &       &       &       &       &       &       &       \\
Fretful       & 0.47 & 1.00  &       &       &       &       &       &       \\
Jealous       & 0.89 & 0.49  & 1.00  &       &       &       &       &       \\
Moody         & 0.54 & 0.57  & 0.56  & 1.00  &       &       &       &       \\
Temperamental & 0.50 & 0.58  & 0.54  & 0.80  & 1.00  &       &       &       \\
Touchy        & 0.61 & 0.66  & 0.64  & 0.79  & 0.91  & 1.00  &       &       \\
Relaxed       & 0.40 & 0.63  & 0.41  & 0.55  & 0.52  & 0.56  & 1.00  &       \\
Unenvious     & 0.83 & 0.49  & 0.80  & 0.57  & 0.52  & 0.59  & 0.45  & 1.00  \\
\bottomrule
\end{tabular}
\end{table}

\subsection{Study 2b}

In Study 2b, we present a parametric approach for developing AI-Agents using sample statistics derived from existing personality data. This method involves extracting key parameters from empirical data, simulating item responses based on these parameters, and then assigning these simulated responses to AI-Agents. By doing so, we aim to provide an efficient alternative or precursor to traditional empirical data collection, facilitating the creation of diverse sets of Agents while maintaining psychometric validity. 
\clearpage

\subsubsection{Methods and Data Analyses}

We generated synthetic BFI-2 data based on sample statistics obtained from  \textcite{soto_next_2017} empirical sample. To do so, we first extracted key statistics from each personality domain. Specifically, within each domain,  we extracted (1) the means and standard deviations for each of the three facets, (2) the $3\times3$  correlation matrix among the three facets, and (3) average correlation among the four items within each facet (See Supplementary Materials for Sample Code).

Using these sample statistics, we then simulated $n=200$ participants' underlying Big Five scores under the following assumptions: (1) personality domains were uncorrelated with each other (i.e., orthogonal domains), (2) within each domain, facets were correlated with each other, (3) within each facet, the items were normally distributed. 

For the BFI-2-Likert and BFI-2-Expanded prompt conditions, the simulated scores were discretized into five categories corresponding to the BFI-2 response scale. These values were then used to generate Likert- or Expanded-formatted prompts for the LLMs, producing AI-Agents with personalities (see Table \ref{tab:examplepromtps}). For the Simple- and Elaborated-Binary conditions, scores were dichotomized and translated into prompts specifying whether the AI-Agents should be high or low on each trait (see Table \ref{tab:examplepromtps}). The statistical simulation script is available in the Supplementary Material.

After creating AI-Agents with personalities, we prompted the AI-Agents to complete the Mini-Markers test just like in Study 2a. With AI-Agents' responses on the Mini-Markers test, we conducted analyses parallel to some of the analyses in Study 2a: 1) correlational analyses between the simulated BFI-2 input scores and the Mini-Markers scores, 2) Omega reliability coefficients, and 3) CFA.

\subsubsection{Results and Discussion}
{\singlespace
\begin{table}[!htbp]
\centering
\caption{Correlation Between BFI-2 Input and LLM's Mini-Marker Test Scores (Study 2b)}
\label{tab:study2b-corr}

\begingroup
\setlength{\tabcolsep}{5pt}
\renewcommand{\arraystretch}{1.05}

\begin{adjustbox}{max width=\textwidth}
\begin{tabular}{@{}ll*{6}{c}@{}}
\toprule
Condition & Model & O & C & E & A & N & Avg \\
\midrule

Human & Human & .75 & .84 & .88 & .80 & .74 & .80 \\

Simple Binary & GPT-3.5 &
  .41/\shadesecond{.73} &
  .46/.57 &
  .60/\shadesecond{.87} &
  .10/.30\textsuperscript{$\dagger$} &
  .46/.54 &
  .41/.60 \\
Simple Binary & GPT-4   &
  .46/.89 &
  .61/\shadesecond{.78} &
  .65/.97 &
  .27/.52 &
  \shadefirst{.76}/.89 &
  .55/\shadesecond{.81} \\
Simple Binary & GPT-4o  &
  .48/.90 &
  .62/\shadesecond{.77} &
  .65/.97 &
  .23/.50 &
  \shadefirst{.75}/.90 &
  .55/\shadesecond{.81} \\
Simple Binary & Llama   &
  .46/.88 &
  .62/\shadesecond{.78} &
  .66/.98 &
  .18/.41 &
  \shadefirst{.73}/.87 &
  .53/\shadesecond{.78} \\
Simple Binary & Deepseek&
  .45/\shadesecond{.80} &
  .58/\shadesecond{.77} &
  .66/.97 &
  .26/.45 &
  \shadefirst{.75}/.89 &
  .54/\shadesecond{.77} \\

Elaborated Binary & GPT-3.5 &
  .34/\shadesecond{.76} &
  .50/.67 &
  .65/.95 &
  .20/.31 &
  .49/.59 &
  .43/.66 \\
Elaborated Binary & GPT-4   &
  .45/.89 &
  .62/\shadesecond{.79} &
  .66/.97 &
  .28/.55 &
  \shadefirst{.75}/.89 &
  .55/\shadesecond{.82} \\
Elaborated Binary & GPT-4o  &
  .47/.91 &
  .61/\shadesecond{.78} &
  .65/.97 &
  .28/.50 &
  \shadefirst{.77}/.91 &
  .56/\shadesecond{.81} \\
Elaborated Binary & Llama   &
  .47/.91 &
  .59/\shadesecond{.77} &
  .65/.98 &
  .23/.47 &
  \shadefirst{.73}/.87 &
  .53/\shadesecond{.80} \\
Elaborated Binary & Deepseek&
  .50/.85 &
  .61/\shadesecond{.77} &
  .66/.97 &
  .23/.51 &
  \shadefirst{.75}/.89 &
  .55/\shadesecond{.80} \\

BFI-2-Likert & GPT-3.5 &
  .51 &
  \shadefirst{.85} &
  \shadefirst{.86} &
  .64 &
  \shadefirst{.71} &
  \shadefirst{.71} \\
BFI-2-Likert & GPT-4   &
  \shadefirst{.68} &
  \shadefirst{.88} &
  \shadefirst{.92} &
  .90 &
  \shadefirst{.81} &
  \shadefirst{.84} \\
BFI-2-Likert & GPT-4o  &
  .87 &
  .92 &
  \shadefirst{.93} &
  .88 &
  .91 &
  .90 \\
BFI-2-Likert & Llama   &
  .92 &
  .90 &
  .93 &
  \shadefirst{.86} &
  .86 &
  .89 \\
BFI-2-Likert & Deepseek&
  \shadefirst{.82} &
  .92 &
  .94 &
  \shadefirst{.87} &
  .85 &
  .88 \\

BFI-2-Expanded & GPT-3.5 &
  \shadefirst{.83} &
  .71 &
  \shadefirst{.87} &
  \shadefirst{.69} &
  \shadefirst{.72} &
  \shadefirst{.77} \\
BFI-2-Expanded & GPT-4   &
  \shadefirst{.84} &
  \shadefirst{.84} &
  \shadefirst{.91} &
  \shadefirst{.82} &
  \shadefirst{.83} &
  \shadefirst{.85} \\
BFI-2-Expanded & GPT-4o  &
  .87 &
  .92 &
  \shadefirst{.92} &
  .90 &
  .90 &
  .90 \\
BFI-2-Expanded & Llama   &
  .88 &
  .92 &
  \shadefirst{.92} &
  \shadefirst{.87} &
  .88 &
  .89 \\
BFI-2-Expanded & Deepseek&
  .86 &
  .92 &
  \shadefirst{.90} &
  \shadefirst{.84} &
  .90 &
  .88 \\
\bottomrule
\end{tabular}
\end{adjustbox}
\endgroup

\vspace{0.5ex}
\begin{minipage}{\textwidth}\footnotesize
\textit{Note.} The sample size for Human and AI-Agents' conditions are different (Human: \textit{N}=438, AI-Agents: \textit{N}=200).
For Simple- and Elaborated-Binary conditions, there are two sets of correlations: (1) correlations between binary Big Five input scores and Mini-Markers output scores (right-hand values), and (2) correlations between original BFI-2 Likert scores and Mini-Markers output scores (left-hand values).
For BFI-2-Likert and -Expanded conditions, correlations were between BFI-2 input and Mini-Markers output.
All correlation coefficients were significantly different from zero at $\alpha=.05$, except where the dagger ($\dagger$) indicates. Light-gray shading marks values \emph{not} statistically different from Human reference after Bonferroni correction across domains ($\alpha=.05$; two-sided Fisher $z$ test on Fisher-transformed $r$).
\end{minipage}

\end{table}
}

\paragraph{Correlational Analysis Between BFI-2 and Mini-Markers}
Study 2b replicated the patterns observed in Study 2a, with slightly lower correlation coefficients (see Table~\ref{tab:study2b-corr}). For instance, for GPT-4o, the average correlations in Study 2a versus Study 2b were .55 vs.\ .59, .56 vs.\ .58, .90 vs.\ .91, and .90 vs.\ .90 for the Simple-Binary, Elaborated-Binary, BFI-2-Likert, and BFI-2-Expanded conditions, respectively. Similar patterns hold at the domain level. Consistent with Study 2a, correlations between Mini-Markers responses and the dichotomized binary BFI-2 input scores were uniformly higher than their corresponding correlations with the original BFI-2 scores. Additionally, the two more complex prompt conditions (BFI-2-Likert and BFI-2-Expanded), which performed at near-identical levels, consistently yielded higher correlations than the two baseline conditions (Simple-Binary and Elaborated-Binary).

In terms of human-AI alignment, Study 2b shows patterns very similar to those in Study 2a. Both BFI-2-Likert and BFI-2-Expanded showed good alignment, with GPT-3.5 and GPT-4 being the best-performing LLMs. For the Simple- and Elaborated-Binary conditions, the correlations between the original BFI-2 and Mini-Markers scores show very poor alignment with humans, except for Neuroticism; meanwhile, correlations between the binary Big Five input and Mini-Markers output scores showed good alignment for Openness and Extraversion but were inflated relative to human correlations for other domains.

\paragraph{Factor Analysis}
Once again, the Study 2b factor analyses showed the same pattern of results as Study 2a\footnote{Full results from factor analyses are provided in the Supplementary Materials}. Consistent with Study 2a, with the exception of GPT-3.5, LLMs exhibited unrealistically good fit and reliability under the Simple- and Elaborated-Binary prompt conditions, and poorer fit and lower reliability under the BFI-2-Likert and BFI-2-Expanded conditions. For example, GPT-4o obtained SRMR values of .02 (Simple-Binary), .02 (Elaborated-Binary), .11 (BFI-2-Likert), and .12 (BFI-2-Expanded), compared to a human reference of .09; and reliability coefficients of .99, .98, .89, and .84, compared to .83 for human reference. As documented in Study 2a, the binary prompt conditions yielded spuriously strong fit because newer LLMs closely track the low/high signals embedded in the prompts and consequently produce highly similar (often near-identical) responses across items within a domain, inflating inter-item correlations and, in turn, model-fit indices and reliability far beyond human levels. By contrast, the Likert and Expanded conditions preserved graded, continuous variation and subtler covariation among items, leading to loadings, fit indices, and reliability that more closely resemble human data (see Tables \ref{tab:study2aloadings} and \ref{tab:factor-analysis}).

Overall, as shown by the results of item-level analyses, input-output correlational analyses and factor analyses, the strong parallelism between Study 2a and Study 2b suggests that creating AI-Agents using sample statistics provides a feasible alternative or precursor to collecting new human data. This method offers researchers a practical tool for exploring personality dynamics and interactions in a controlled, scalable environment, while maintaining psychometric validity comparable to traditional human-subject research.

\begin{table}[htbp]
\centering
\setlength{\tabcolsep}{1.5pt}
\renewcommand{\arraystretch}{1.2}
\tiny

\begin{adjustbox}{angle=-90, center, max width=\textheight}
\begin{minipage}{\textheight}
\centering
\caption{Fit Measures for CFA Models in Study 2a}
\label{tab:factor-analysis}
\begin{tabular}{ll c ccccc ccccc ccccc ccccc}
\toprule
\multirow{2}{*}{\textbf{Fit Measure}} & \multirow{2}{*}{} &
\multicolumn{1}{c}{\textbf{Human}} &
\multicolumn{5}{c}{\textbf{Simple-Binary}} &
\multicolumn{5}{c}{\textbf{Elaborated-Binary}} &
\multicolumn{5}{c}{\textbf{BFI-2-Likert}} &
\multicolumn{5}{c}{\textbf{BFI-2-Expanded}} \\
\cmidrule(lr){3-3}\cmidrule(lr){4-8}\cmidrule(lr){9-13}\cmidrule(lr){14-18}\cmidrule(lr){19-23}
& & Ref & 3.5 & 4 & 4o & Llama & DS & 3.5 & 4 & 4o & Llama & DS & 3.5 & 4 & 4o & Llama & DS & 3.5 & 4 & 4o & Llama & DS \\
\midrule
Extraversion & Chi-Squared & 238.59 & 587.55 & 323.25 & 833.66 & 741.86 & 163.11 & 562.79 & 211.50 & 664.82 & 728.75 & 144.12 & 146.38 & 395.06 & 668.01 & 558.96 & 596.41 & 754.33 & 471.32 & 906.42 & 433.74 & 444.58 \\
 & CFI & .87 & \cellcolor{LightShade}.93 & \cellcolor{DarkShade}.97 & \cellcolor{LightShade}.93 & \cellcolor{DarkShade}.95 & .88 & \cellcolor{LightShade}.93 & \cellcolor{DarkShade}.98 & \cellcolor{LightShade}.94 & \cellcolor{DarkShade}.95 & .89 & .38 & .87 & .86 & .87 & .85 & .84 & .87 & .82 & \cellcolor{LightShade}.92 & \cellcolor{LightShade}.90 \\
 & RMSEA & .16 & .26 & .19 & .30 & .29 & .43 & .25 & .15 & .27 & .28 & .40 & .41 & .21 & .27 & .25 & .26 & .29 & .23 & .32 & .22 & .22 \\
 & SRMR & \cellcolor{LightShade}.07 & \cellcolor{DarkShade}.03 & \cellcolor{DarkShade}.01 & \cellcolor{DarkShade}.01 & \cellcolor{DarkShade}.00 & \cellcolor{DarkShade}.01 & \cellcolor{DarkShade}.03 & \cellcolor{DarkShade}.00 & \cellcolor{DarkShade}.01 & \cellcolor{DarkShade}.00 & \cellcolor{DarkShade}.01 & .24 & .10 & \cellcolor{LightShade}.07 & \cellcolor{LightShade}.07 & .10 & .10 & .09 & \cellcolor{DarkShade}.05 & \cellcolor{DarkShade}.03 & \cellcolor{DarkShade}.04 \\
Agreeableness & Chi-Squared & 225.69 & 562.12 & 333.68 & 975.38 & 3244.27 & 2822.90 & 572.83 & 381.25 & 1082.12 & 4006.49 & 2627.98 & 1236.72 & 578.75 & 873.35 & 1232.21 & 754.11 & 895.40 & 682.44 & 695.87 & 1209.57 & 1211.47 \\
 & CFI & .84 & .80 & \cellcolor{DarkShade}.95 & .84 & .60 & .70 & .79 & \cellcolor{LightShade}.94 & .83 & .53 & .71 & .63 & .81 & .78 & .74 & .79 & .73 & .76 & .82 & .72 & .69 \\
 & RMSEA & .15 & .25 & .19 & .33 & .61 & .57 & .25 & .20 & .35 & .68 & .55 & .37 & .25 & .31 & .37 & .29 & .32 & .28 & .28 & .37 & .37 \\
 & SRMR & \cellcolor{LightShade}.08 & .17 & \cellcolor{DarkShade}.03 & \cellcolor{LightShade}.06 & .14 & .12 & .19 & \cellcolor{DarkShade}.03 & \cellcolor{LightShade}.06 & .13 & .14 & .29 & .11 & \cellcolor{LightShade}.08 & .10 & .09 & .18 & .09 & \cellcolor{LightShade}.07 & \cellcolor{LightShade}.08 & .09 \\
Conscientiousness & Chi-Squared & 317.88 & 1147.92 & 202.22 & 357.77 & 1409.29 & 1356.58 & 1252.59 & 154.42 & 453.51 & 1388.33 & 877.57 & 767.39 & 631.95 & 626.56 & 723.88 & 759.83 & 916.01 & 666.71 & 536.44 & 915.88 & 901.91 \\
 & CFI & .79 & .71 & \cellcolor{DarkShade}.98 & \cellcolor{DarkShade}.96 & .85 & .85 & .67 & \cellcolor{DarkShade}.98 & \cellcolor{DarkShade}.95 & .85 & \cellcolor{LightShade}.90 & .62 & .69 & .80 & .79 & .70 & .71 & .72 & .84 & .79 & .74 \\
 & RMSEA & .18 & .36 & .14 & .20 & .40 & .39 & .38 & .12 & .22 & .40 & .31 & .29 & .26 & .26 & .28 & .29 & .32 & .27 & .24 & .32 & .32 \\
 & SRMR & .09 & .14 & \cellcolor{DarkShade}.01 & \cellcolor{DarkShade}.01 & \cellcolor{DarkShade}.03 & \cellcolor{DarkShade}.03 & .17 & \cellcolor{DarkShade}.01 & \cellcolor{DarkShade}.02 & \cellcolor{DarkShade}.03 & \cellcolor{DarkShade}.02 & .20 & .14 & \cellcolor{LightShade}.08 & \cellcolor{LightShade}.08 & .11 & .10 & .11 & \cellcolor{LightShade}.07 & \cellcolor{LightShade}.08 & .11 \\
Negative Emotionality & Chi-Squared & 329.78 & 1484.58 & 1253.61 & 477.87 & 1628.75 & 1384.60 & 1494.83 & 1273.36 & 656.71 & 1818.51 & 1083.63 & 775.23 & 798.21 & 100.01 & 1176.54 & 1062.65 & 1646.45 & 966.10 & 1122.56 & 1656.90 & 1327.83 \\
 & CFI & .78 & .36 & .83 & \cellcolor{DarkShade}.96 & .78 & .86 & .38 & .83 & \cellcolor{LightShade}.94 & .76 & .88 & .75 & .61 & .70 & .70 & .72 & .42 & .58 & .72 & .59 & .67 \\
 & RMSEA & .19 & .41 & .38 & .23 & .43 & .40 & .41 & .38 & .27 & .45 & .35 & .29 & .30 & .33 & .36 & .34 & .43 & .33 & .35 & .43 & .39 \\
 & SRMR & .10 & .28 & .14 & \cellcolor{DarkShade}.01 & .12 & \cellcolor{DarkShade}.03 & .29 & .13 & \cellcolor{DarkShade}.01 & .12 & \cellcolor{DarkShade}.04 & .20 & .17 & .12 & .11 & .19 & .32 & .24 & .13 & .21 & .21 \\
Open-Mindedness & Chi-Squared & 418.21 & 449.04 & 365.43 & 791.24 & 1513.55 & 1754.92 & 443.68 & 344.24 & 752.06 & 1596.25 & 120.19 & 846.06 & 152.88 & 1659.83 & 1914.50 & 2266.87 & 1551.01 & 2458.43 & 2189.52 & 2053.83 & 2841.07 \\
 & CFI & .71 & .83 & \cellcolor{LightShade}.93 & .89 & .75 & .77 & .85 & \cellcolor{LightShade}.94 & \cellcolor{LightShade}.90 & .74 & .83 & .66 & .58 & .59 & .60 & .53 & .50 & .51 & .54 & .54 & .54 \\
 & RMSEA & .21 & .22 & .20 & .30 & .41 & .45 & .22 & .19 & .29 & .42 & .37 & .31 & .41 & .43 & .47 & .51 & .42 & .53 & .50 & .48 & .57 \\
 & SRMR & .12 & .12 & \cellcolor{DarkShade}.03 & \cellcolor{DarkShade}.03 & .09 & \cellcolor{LightShade}.07 & .11 & \cellcolor{DarkShade}.03 & \cellcolor{DarkShade}.03 & .09 & \cellcolor{DarkShade}.05 & .13 & .28 & .14 & .13 & .14 & .18 & .24 & .22 & .20 & .26 \\
\bottomrule
\end{tabular}

\vspace{1em}
\raggedright
\scriptsize
\textbf{Note.} Column blocks list models as: 3.5 = GPT-3.5 Turbo (0125); 4 = GPT-4; 4o = GPT-4o; Llama = Llama-3.30-70B-Instruct; DS = DeepSeek-V3. ``Human'' is the reference fit; ``Ref'' is its summary column. Averages are computed across the five subscales within each condition model. Values rounded to 2 decimals (leading zero omitted). The asterisk beside the chi-square for Open-Mindedness $\times$ GPT-4 $\times$ BFI-2-Likert indicates that the model fit yielded a warning (Heywood case). Light gray = acceptable fit (CFI $\ge$ .90, RMSEA $\le$ .08, SRMR $\le$ .08); dark gray = very good fit (CFI $\ge$ .95, RMSEA $\le$ .05, SRMR $\le$ .05).
\end{minipage}
\end{adjustbox}
\end{table}

\section{Study 3: Further Validating AI-Agents with Real-Life Moral and Risk-Taking Vignettes}
In Study 3, we aim to further validate AI-Agents with real-like moral and risk-taking vignettes and examine their alignment with human behaviors. 
This investigation seeks to elucidate both the potential and limitations of employing AI-Agents in behavioral science studies.

\subsection{Methods}
\subsubsection{Measurement}
In this study, we crafted ten vignettes: five risk-taking and five moral dilemma vignettes. The five risk-taking vignettes were designed to test an individual's risk-taking tendency versus risk-avoidance tendency. The specific vignettes included: 1) embarking on an entrepreneurial venture, 2) making significant investments, 3) confessing romantic feelings to a close friend, 4) participating in extreme sports, and 5) opting to study overseas (See Appendix II).
For the risk-taking vignettes, higher scores indicate higher risk-seeking and lower risk-avoidance.

The five moral dilemma vignettes were designed to assess individuals' empathetic tendency versus rule adherence tendency. Given that the majority of currently available LLMs have been safety-aligned and instruction fine-tuned \parencite{biedma2024beyond}, which typically skews them away from engaging in severe moral judgments (e.g., the Trolley Problem), we introduced a series of everyday moral dilemmas. These dilemmas necessitated choosing between upholding a standard and prioritizing empathy. Examples include decisions on 1) reporting a friend for cheating on a quiz, 2) addressing a colleague's misappropriation of office supplies, 3) underage drinking, 4) disclosing confidential information that could save lives, and 5) providing candid feedback on subpar performance (see Appendix III). 
For the moral dilemmas vignettes, higher scores indicate higher rule adherence tendency and lower empathetic tendency.

\subsubsection{Procedure}

For human data, we conducted a survey study at a Canadian University. 
Participants were asked to complete the BFI-2 Expanded-format scale and respond to the ten vignettes. After applying exclusion criteria based on English proficiency, consent for data deposit, and survey completion, we retained a sample of 276 participants. The mean age of participants was 19.65 years ($SD$ = 3.88). Regarding gender identity, the majority of participants identified as female (80.4\%), with 15.6\% identifying as male and 4\% identified as Other. The ethnic composition of the sample was diverse: 26.8\% as European/Caucasian, 25.0\% as South Asian, 7.3\% as African, 6.9\% as East Asian, 2.5\% as Latino and Hispanic and 28.3\% identified as Other.

To create the AI-Agents, same as Study 2, first, we used the four prompting strategies (Simple-Binary, Elaborated-Binary, BFI-2-Likert and BFI-2-Expanded, see Table \ref{tab:examplepromtps}) and five LLMs (GPT-3.5, GPT-4, GPT-4o, Llama, and DeepSeek).  For each prompt $\times$ LLM condition, we created 276 AI-Agents, each corresponding to a human participant's personality. We then prompted the AI-Agents to read five risk-taking vignettes and five moral dilemmas and to indicate their decisions on a 1-10 scale (see Appendices C and D for prompts).

\subsubsection{Data Analyses}

Similar to Study 2, in Study 3, we conducted analyses comparing human and AI-Agents responses on the ten vignettes. We compared the means and distributions of their responses using paired-wise $t$-test and KS tests, respectively. In addition, we conducted regression analyses in which the original Big Five trait scores assigned to the AI-Agents were used to predict their responses to the ten vignettes. All significance tests employed Bonferroni corrections to control the family-wise Type I error rate.

\subsubsection{Results and Discussion}

\begin{figure}[htbp]
    \caption{Selected Scores on Vignettes For Human and AI-Agents}
    \label{fig:vingettes}
    \includegraphics[width=0.9\textwidth]{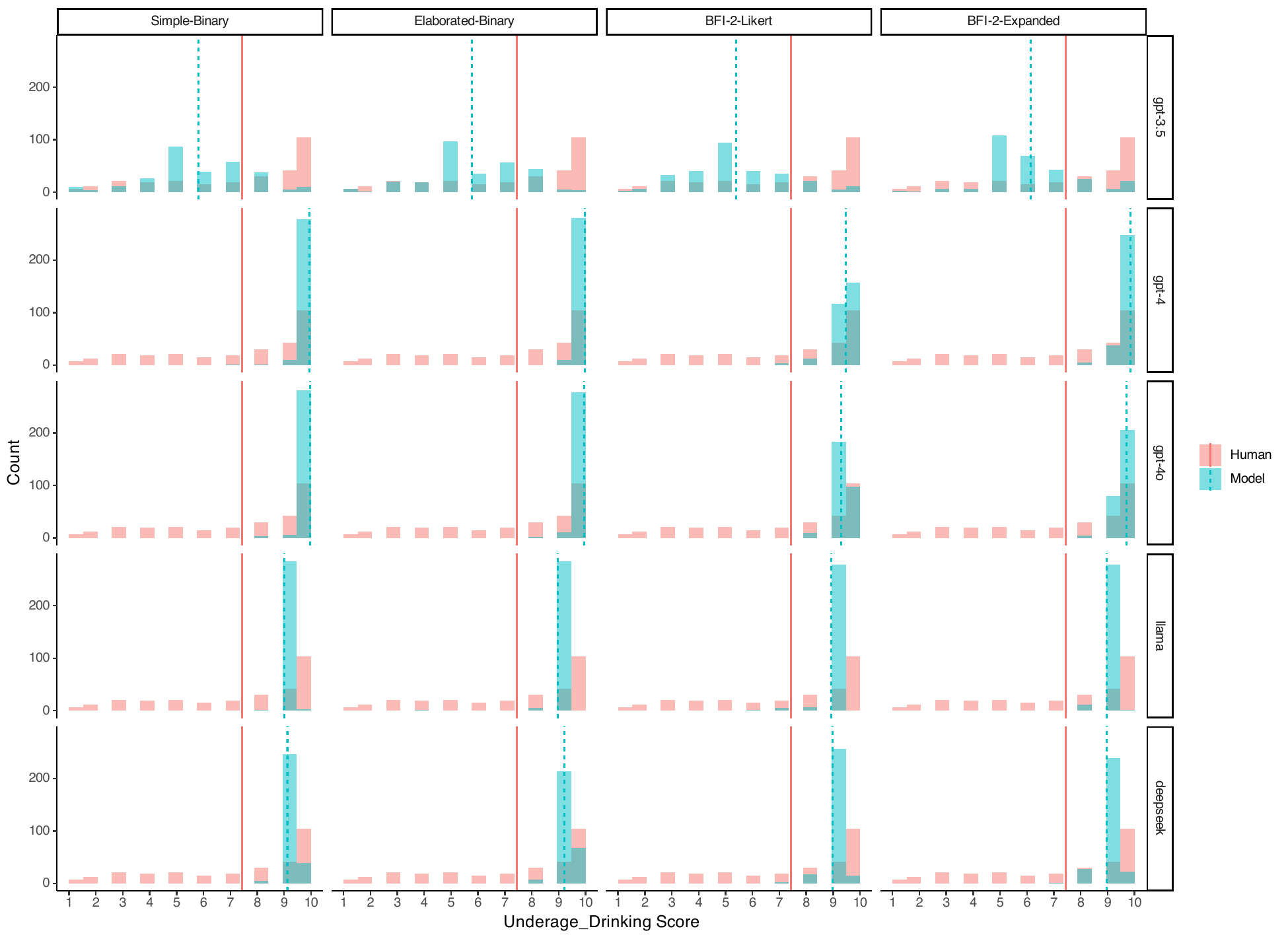}
     \includegraphics[width=0.9\textwidth]{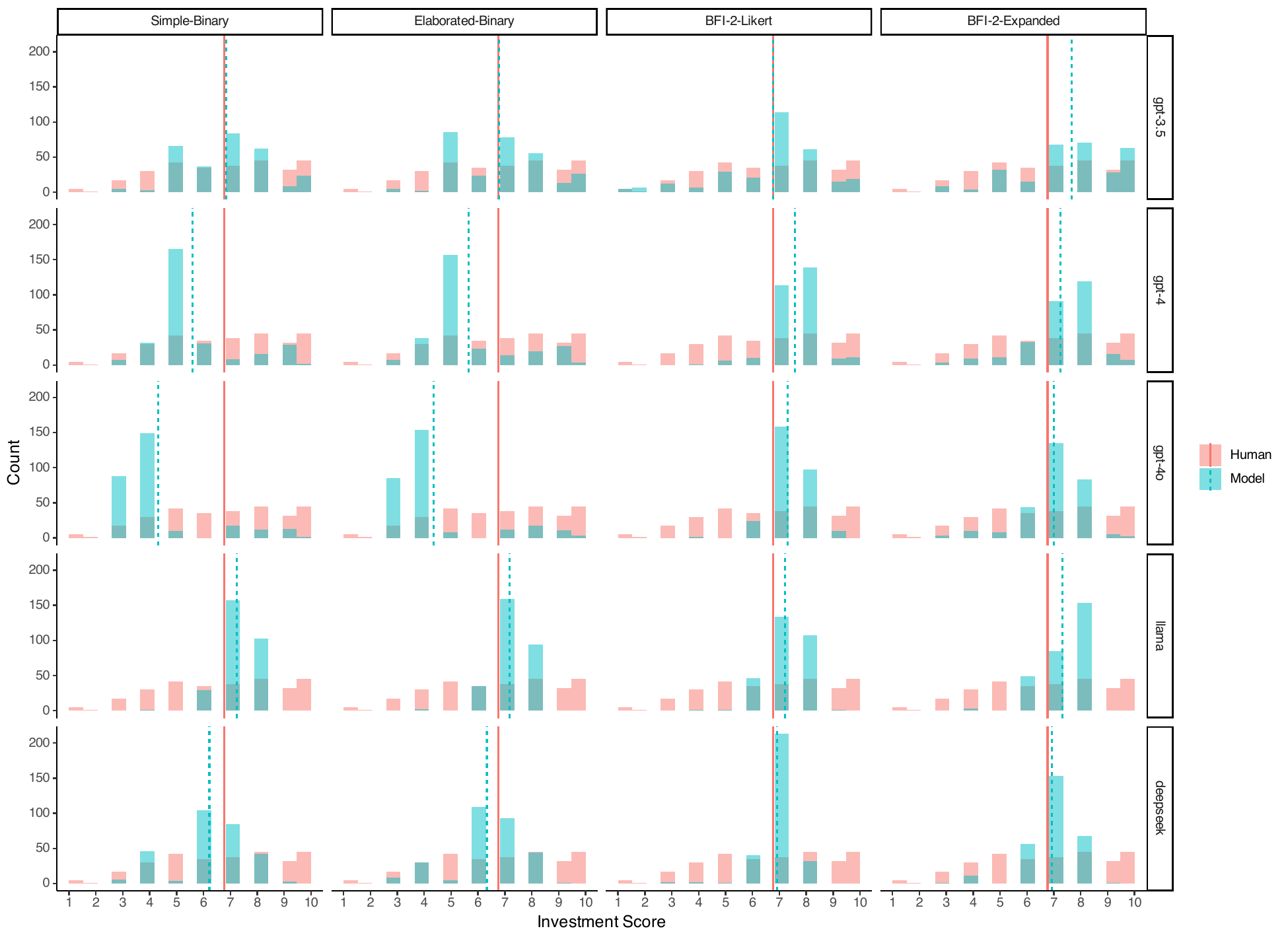}

\footnotesize
\emph{Note.} The red line indicates the mean of the human responses. The dotted blue lines indicate the means of AI-Agents across conditions.
\end{figure}

\paragraph{Comparing Human and AI-Agents Responses on the Risk and Moral Vignettes} 
Tables \ref{tab:mean_diffs_moral_risk_booktabs} and \ref{tab:ks_stats_moral_risk_booktabs} show the mean and distribution differences between AI-Agents and human responses for the ten vignettes. Figure \ref{fig:vingettes} shows graphic representations of these differences in selected conditions. 

The most noticeable pattern of results is that the newer LLMs are clearly fine-tuned to provide morally acceptable responses in the moral dilemma vignettes. As shown in Table \ref{tab:mean_diffs_moral_risk_booktabs} and Figure \ref{fig:vingettes}, for dilemmas involving confidential information, underage drinking, and exam cheating, AI-Agents created with newer LLMs consistently scored substantially higher across prompt conditions, with differences ranging from 1.02 to 3.88. As a result, the AI-Agents' score distributions are heavily skewed and significantly different from those of humans (see Table \ref{tab:ks_stats_moral_risk_booktabs} and Figure \ref{fig:vingettes}). In contrast, AI-Agents created by the older LLM (GPT-3.5) often scored lower on moral vignettes than human participants. This clear divergence suggests that the newer LLMs have been deliberately fine-tuned for safety, aligning their responses more closely with socially desirable moral norms. This pattern of results is consistent with our expectation that the newer LLMs have been deliberately fine-tuned for safety reasons. 

For risk-taking vignettes, the mean and distributional differences between AI-Agents and human responses were considerably lower, especially in the BFI-2-Likert and -Expanded conditions (see Tables \ref{tab:mean_diffs_moral_risk_booktabs} and \ref{tab:ks_stats_moral_risk_booktabs} and Figure \ref{fig:vingettes}). In fact, for the investment and confession vignettes, AI-Agents produced mean scores that were comparable to those of humans (as indicated by the shaded cells in Table \ref{tab:mean_diffs_moral_risk_booktabs}). However, even in these cases, the responses of AI-Agents, especially those generated by the newer LLMs,displayed substantially less variation than human responses (see Figure \ref{fig:vingettes}).
In fact,  the investment and confession risk-taking vignette had similar mean scores compared to humans (as shown by the shaded cells in Table \ref{tab:mean_diffs_moral_risk_booktabs}), although the AI-Agents' responses still showed considerably less variation than human responses for the newer LLMs (see Figure \ref{fig:vingettes}).

Comparing across the prompt conditions, for the moral dilemma vignettes, all prompt conditions yielded responses that were heavily skewed to the right (see Figure \ref{fig:vingettes}. On the other hand, for risk-taking vignettes, the AI-Agents in the BFI-2-Likert and -Expanded conditions generated more variable responses, whereas those in the Simple- and Elaborated-Binary conditions continued to display skewed distributions.

\begin{table}[htbp]
\scriptsize
\setlength{\tabcolsep}{1.5pt}
\renewcommand{\arraystretch}{1}
\caption{\textit{Mean Differences between AI-Agents and Human Responses on the Risk and Moral Vignettes}}
\label{tab:mean_diffs_moral_risk_booktabs}
\begin{tabular}{llcccccccccc} 
\toprule
\textbf{Condition} & \textbf{LLM} & \textbf{Conf} & \textbf{Under} & \textbf{Exam} & \textbf{Honest} & \textbf{Work} & \textbf{Invest} & \textbf{Ext} & \textbf{Entre} & \textbf{Confess} & \textbf{Study} \\
 & & \textbf{Info} & \textbf{Age} & \textbf{Cheat} & \textbf{FB} & \textbf{Theft} &  & \textbf{Sports} & \textbf{Vent} &  & \textbf{Abroad} \\
\midrule
Simple-Binary & GPT-3.5 & -0.74 & -1.62 & 1.02 & 0.49 & 0.76 & \cellcolor{gray!20}{0.08} & -2.68 & -2.86 & \cellcolor{gray!20}{0.23} & -2.27 \\
Simple-Binary & GPT-4 & 3.88 & 2.51 & 4.98 & 1.12 & 2.09 & -1.16 & 1.50 & -0.99 & -1.70 & -1.10 \\
Simple-Binary & GPT-4o & 2.15 & 2.52 & 3.53 & 0.78 & 3.15 & -2.45 & \cellcolor{gray!20}{0.33} & -1.69 & -2.26 & -1.39 \\
Simple-Binary & Llama & 2.16 & 1.57 & 3.84 & 1.03 & 2.93 & 0.48 & 1.18 & \cellcolor{gray!20}{-0.12} & -1.37 & -0.69 \\
Simple-Binary & DeepSeek & 1.93 & 1.68 & 2.93 & \cellcolor{gray!20}{0.34} & 2.18 & -0.54 & 1.09 & -1.28 & -1.60 & -0.83 \\
Elaborated-Binary & GPT-3.5 & \cellcolor{gray!20}{-0.47} & -1.65 & 1.04 & \cellcolor{gray!20}{0.44} & 0.81 & \cellcolor{gray!20}{0.02} & -2.60 & -2.98 & \cellcolor{gray!20}{0.34} & -2.43 \\
Elaborated-Binary & GPT-4 & 3.84 & 2.53 & 4.84 & 1.03 & 2.02 & -1.12 & 1.49 & -0.92 & -1.64 & -1.11 \\
Elaborated-Binary & GPT-4o & 2.09 & 2.51 & 3.57 & 0.76 & 3.18 & -2.42 & \cellcolor{gray!20}{0.22} & -1.74 & -2.28 & -1.40 \\
Elaborated-Binary & Llama & 2.12 & 1.53 & 3.84 & 1.01 & 2.89 & \cellcolor{gray!20}{0.42} & 1.20 & \cellcolor{gray!20}{-0.10} & -1.33 & -0.71 \\
Elaborated-Binary & DeepSeek & 1.96 & 1.77 & 3.00 & 0.47 & 2.19 & \cellcolor{gray!20}{-0.44} & 1.24 & -1.17 & -1.60 & -0.98 \\
BFI-2-Likert & GPT-3.5 & -1.49 & -2.03 & -0.64 & -0.90 & 0.53 & \cellcolor{gray!20}{0.01} & -2.56 & -3.24 & \cellcolor{gray!20}{-0.39} & -2.13 \\
BFI-2-Likert & GPT-4 & 2.93 & 2.04 & 4.21 & 1.23 & 3.13 & 0.80 & 2.37 & 1.83 & \cellcolor{gray!20}{0.11} & 1.63 \\
BFI-2-Likert & GPT-4o & 1.44 & 1.87 & 3.06 & -0.97 & 2.74 & 0.54 & 2.22 & 1.73 & 0.86 & 1.41 \\
BFI-2-Likert & Llama & 2.09 & 1.50 & 3.86 & 0.85 & 3.21 & 0.43 & 0.94 & 0.66 & \cellcolor{gray!20}{-0.03} & 1.09 \\
BFI-2-Likert & DeepSeek & 2.10 & 1.54 & 2.89 & \cellcolor{gray!20}{0.16} & 2.13 & \cellcolor{gray!20}{0.15} & 1.53 & \cellcolor{gray!20}{-0.03} & -0.68 & 0.66 \\
BFI-2-Expanded & GPT-3.5 & -2.52 & -1.30 & -0.80 & -0.72 & 0.79 & 0.89 & -0.77 & -0.91 & 1.28 & \cellcolor{gray!20}{-0.37} \\
BFI-2-Expanded & GPT-4 & 3.52 & 2.40 & 4.31 & 0.83 & 2.20 & 0.49 & 2.60 & 1.23 & \cellcolor{gray!20}{-0.10} & 0.80 \\
BFI-2-Expanded & GPT-4o & 1.81 & 2.26 & 3.12 & \cellcolor{gray!20}{-0.28} & 2.88 & \cellcolor{gray!20}{0.22} & 2.08 & 0.81 & \cellcolor{gray!20}{-0.06} & \cellcolor{gray!20}{0.17} \\
BFI-2-Expanded & Llama & 2.08 & 1.53 & 3.48 & 0.50 & 3.27 & 0.56 & 0.82 & \cellcolor{gray!20}{0.21} & \cellcolor{gray!20}{-0.44} & 1.51 \\
BFI-2-Expanded & DeepSeek & 2.03 & 1.54 & 2.62 & \cellcolor{gray!20}{0.12} & 2.12 & \cellcolor{gray!20}{0.16} & 1.51 & \cellcolor{gray!20}{-0.23} & -0.68 & 0.55 \\
\bottomrule
\end{tabular}

\begingroup\scriptsize
\smallskip
\emph{Note}: Values are AI $-$ Human mean differences. Unshaded cells indicate significant paired-sample $t$-tests (Bonferroni-corrected); gray cells indicate non-significant results. Conf Info=Vignette about disclosing confidential information; Under Age=Vignette about underage drinking; Exam Cheat=Vignette about exam cheating; Honest FB=Vignette about providing candid feedback on subpar performance; Worth Theft=Vignette about workplace theft; Invest=Vignette about making investment; Ext Sports=Vignette about participating in extreme sports; Entre Vent=Vignette about embarking on an entrepreneurial venture; Confess=Vignette about confessing to a close friend; Study Abroad=Vignette about studying abroad.
\endgroup
\end{table}

\begin{table}[htbp]
\scriptsize
\setlength{\tabcolsep}{1.5pt}
\renewcommand{\arraystretch}{1}
\caption{\textit{Kolmogorov-Smirnov Test Statistics Comparing AI-Agents and Human Responses on the Risk and Moral Vignettes}}
\label{tab:ks_stats_moral_risk_booktabs}
\begin{tabular}{llcccccccccc} 
\toprule
\textbf{Condition} & \textbf{LLM} & \textbf{Conf} & \textbf{Under} & \textbf{Exam} & \textbf{Honest} & \textbf{Work} & \textbf{Invest} & \textbf{Ext} & \textbf{Entre} & \textbf{Confess} & \textbf{Study} \\
 & & \textbf{Info} & \textbf{Age} & \textbf{Cheat} & \textbf{FB} & \textbf{Theft} &  & \textbf{Sports} & \textbf{Vent} &  & \textbf{Abroad} \\
\midrule
Simple-Binary & GPT-3.5 & 0.18 & 0.45 & 0.26 & 0.20 & 0.43 & 0.16 & 0.58 & 0.67 & 0.27 & 0.49 \\
Simple-Binary & GPT-4 & 0.78 & 0.60 & 0.86 & 0.28 & 0.49 & 0.38 & 0.35 & 0.36 & 0.41 & 0.29 \\
Simple-Binary & GPT-4o & 0.54 & 0.61 & 0.79 & 0.40 & 0.74 & 0.63 & 0.29 & 0.48 & 0.44 & 0.39 \\
Simple-Binary & Llama & 0.70 & 0.49 & 0.86 & 0.49 & 0.80 & 0.34 & 0.33 & 0.24 & 0.48 & 0.26 \\
Simple-Binary & DeepSeek & 0.60 & 0.48 & 0.79 & 0.43 & 0.70 & 0.27 & 0.33 & 0.38 & 0.38 & 0.22 \\
Elaborated-Binary & GPT-3.5 & \cellcolor{gray!20}{0.14} & 0.47 & 0.29 & 0.19 & 0.41 & 0.16 & 0.58 & 0.69 & 0.29 & 0.51 \\
Elaborated-Binary & GPT-4 & 0.76 & 0.61 & 0.84 & 0.29 & 0.50 & 0.37 & 0.35 & 0.37 & 0.40 & 0.30 \\
Elaborated-Binary & GPT-4o & 0.53 & 0.60 & 0.80 & 0.40 & 0.72 & 0.64 & 0.30 & 0.50 & 0.46 & 0.40 \\
Elaborated-Binary & Llama & 0.69 & 0.48 & 0.87 & 0.49 & 0.80 & 0.32 & 0.35 & 0.24 & 0.45 & 0.27 \\
Elaborated-Binary & DeepSeek & 0.59 & 0.47 & 0.76 & 0.42 & 0.69 & 0.27 & 0.36 & 0.37 & 0.40 & 0.23 \\
BFI-2-Likert & GPT-3.5 & 0.28 & 0.48 & 0.22 & 0.28 & 0.32 & 0.17 & 0.49 & 0.73 & 0.26 & 0.48 \\
BFI-2-Likert & GPT-4 & 0.60 & 0.44 & 0.76 & 0.25 & 0.57 & 0.39 & 0.50 & 0.50 & 0.32 & 0.47 \\
BFI-2-Likert & GPT-4o & 0.44 & 0.46 & 0.72 & 0.41 & 0.68 & 0.36 & 0.52 & 0.44 & 0.32 & 0.47 \\
BFI-2-Likert & Llama & 0.66 & 0.46 & 0.87 & 0.43 & 0.85 & 0.32 & 0.27 & 0.30 & 0.33 & 0.47 \\
BFI-2-Likert & DeepSeek & 0.62 & 0.43 & 0.74 & 0.37 & 0.69 & 0.31 & 0.48 & 0.30 & 0.34 & 0.40 \\
BFI-2-Expanded & GPT-3.5 & 0.50 & 0.42 & 0.22 & 0.24 & 0.33 & 0.24 & 0.20 & 0.40 & 0.31 & 0.25 \\
BFI-2-Expanded & GPT-4 & 0.71 & 0.50 & 0.83 & 0.24 & 0.52 & 0.25 & 0.58 & 0.24 & 0.26 & 0.33 \\
BFI-2-Expanded & GPT-4o & 0.46 & 0.48 & 0.72 & 0.31 & 0.71 & 0.26 & 0.47 & 0.24 & 0.21 & 0.30 \\
BFI-2-Expanded & Llama & 0.67 & 0.46 & 0.86 & 0.40 & 0.74 & 0.32 & 0.24 & 0.22 & 0.28 & 0.47 \\
BFI-2-Expanded & DeepSeek & 0.54 & 0.40 & 0.74 & 0.26 & 0.62 & 0.29 & 0.41 & 0.24 & 0.33 & 0.33 \\
\bottomrule
\end{tabular}

\begingroup\scriptsize
\medskip
\emph{Note}: Values are KS $D$ statistics. Unshaded cells indicate significant KS tests (Bonferroni-corrected); gray cells indicate non-significant results. Conf Info=Vignette about disclosing confidential information; Under Age=Vignette about underage drinking; Exam Cheat=Vignette about exam cheating; Honest FB=Vignette about providing candid feedback on subpar performance; Worth Theft=Vignette about workplace theft; Invest=Vignette about making investment; Ext Sports=Vignette about participating in extreme sports; Entre Vent=Vignette about embarking on an entrepreneurial venture; Confess=Vignette about confessing to a close friend; Study Abroad=Vignette about studying abroad.\\
\endgroup
\end{table}

\normalsize

\paragraph{Predicting Risk and Moral Vignettes Scores Based on Big Five}

Table \ref{tab:combined_coefficients} of standardized regression coefficients shows how each Big Five domain predicts moral- and risk-taking decisions across response formats and models. Among human participants, for the risk-taking vignettes, individuals higher in Openness and Extraversion were significantly more likely to endorse risky actions (O: $\beta$= .14*, E: $\beta$= .23*), whereas higher Neuroticism predicted a greater avoidance of risk ($\beta$= –.15*). Conscientiousness and Agreeableness show no significant associations with risk. For the moral dilemma vignettes, Conscientiousness positively predicted prosocial choices ($\beta$= .21***), while Neuroticism negatively predicted them ($\beta$= –.15*); the remaining domains were not significant. These human patterns provided a reference for interpreting AI-Agent behavior. 

\paragraph{\textit{Risk-Taking Vignettes Scores}} For risk-taking, AI-Agents created with Simple- and Elaborated-Binary and BFI-2-Expanded were all good at recovering human prediction patterns (i.e., direction and significance). Across simple- and elaborated-binary and the BFI-2-Expanded prompt condition, AI-Agents displayed a common pattern that mirrored human risk-taking behavior. In all three conditions, coefficients for Openness and Extraversion were large and positive, while Neuroticism was strongly negative. These effects were much larger than those observed in humans and remained significant across most models. It suggests that AI-Agents consistently linked curiosity and sociability with greater risk appetite and associate emotional instability with risk aversion. 

Meanwhile, the BFI-2-Expanded prompt condition differed from two baseline conditions: the expanded-prompt elicited larger Extraversion coefficients ($\beta$ =0.65-0.81) and Neuroticism coefficients ($\beta$= .-29-.-41) but Openness coefficients that were slightly smaller or comparable to those in the baseline (expanded $\beta$ =.19-.35 vs. baseline $\beta$ =.26-.41), and it attenuated inconsistent and spurious effects of other domains. In contrast, the baseline binary prompts sometimes produced small but significant positive Agreeableness and Conscientiousness coefficients (e.g., GPT-4 A=0.20*** in simple-binary; GPT-4 C=0.15*). Thus, both prompting approaches captured the human-like pattern of risk-taking, but the expanded-prompt condition provided a cleaner representation; the main exception was that several GPT models continue to show modest positive effects for Agreeableness in the baseline conditions. 

In contrast, under the BFI-2-Likert condition, coefficient patterns diverged from the human reference. Only isolated recoveries appeared: Openness was positive only for GPT-3.5 ($\beta = .13^{*}$),  Extraversion only for Llama ($\beta = .31^{*}$), 
and the negative Neuroticism effect for GPT-3.5 ($\beta = - .13^{*}$), 
GPT-4o ($\beta = - .13^{*}$), and Llama ($\beta = - .20^{***}$). Signs and magnitudes for the remaining domains frequently flipped or attenuated, and this condition did not reliably reproduce the human non-significant associations for Conscientiousness and Agreeableness. Overall, unlike the BFI-2-Expanded and binary prompt conditions, the Likert condition failed to recover most of the associations between personality and risk-taking behaviors.

\paragraph{\textit{Moral Dilemma Scores}} 
Patterns for moral dilemma are similar to risk-taking patterns. In binary and expanded prompt conditions, AI-Agents' decisions largely followed the human pattern: prosocial choices increased with higher Conscientiousness and decreased with higher Neuroticism, whereas the other domains showed little effect. Expanded prompt conditions outperformed the two binary prompt conditions, as they more consistently reproduced the significant positive coefficients for Conscientiousness and significant positive coefficients for Neuroticism (except for GPT-3.5 and GPT-4) while keeping the non-significant coefficients for the other three domains. These results show that the expanded-prompt condition produced clearer and stronger effects while attenuating spurious effects and yielding a closer match to human data.

In contrast, the BFI-2-Likert prompt condition shows much worse alignment with the human reference. It only fully recovered the non-significant coefficients for Openesses, and partially recovers the significant positive coefficients for Conscientiousness (GPT-3.5: $\beta$=.18**, GPT-4: $\beta$=.23***, and Deepseek: $\beta$=.17**), while the rest of domains show inconsistent patterns that fail to match the human reference. 

\paragraph{\textit{Model comparisons across Risk-taking and Moral Dilemma vignettes}} 
Across both risk-taking and moral dilemma vignettes, model performance varied dramatically between the BFI-2-Likert and BFI-2-Expanded conditions. In the BFI-2-Expanded prompt condition, GPT-4o demonstrated the best alignment with human reference, followed by GPT-4 and Llama, showing the most robust reproduction of human patterns with consistently large positive coefficients for Openness (risk: $\beta$ =.32***) and Extraversion (risk: $\beta$=.81***; moral: non-significant as expected), and strong negative Neuroticism effects across both vignette types. 

The BFI-2-Likert condition revealed striking model differences and generally poor performance. Most concerningly, GPT-4 and GPT-4o showed patterns directly opposite to human reference in this format, producing negative coefficients where humans showed positive ones (e.g., GPT-4o moral Extraversion: $\beta$=-.21*** vs. human non-significant). GPT-3.5 and Llama maintained better directional consistency in the Likert format, with GPT-3.5 partially recovering human risk patterns (Openness: $\beta$=.13*) and Llama showing robust Extraversion effects (risk: $\beta$=.31***). DeepSeek showed intermediate performance, with some preserved directional relationships but inconsistent magnitudes.

Overall, newer, more sophisticated models (GPT-4, GPT-4o) showed better format sensitivity, performing very well in the BFI-2-Expanded condition but poorly in the BFI-2-Likert condition. This suggests that advanced models may be more susceptible to prompt-specific linguistic cues. Meanwhile, in BFI-2-Expanded condition, AI-Agent consistently showed amplified true effects while minimizing spurious associations across all models, indicating its superior validity for simulating personality-behavior associations.

One plausible account of the BFI-2-Likert format’s weaker behavioral prediction, even relative to the binary baselines, is its indirectness. In the Likert condition, trait information is conveyed numerically, so the model must first translate numbers into semantic descriptions before mapping those meanings onto behavioral choices, introducing an extra decoding step. By contrast, both BFI-2-Expanded and the binary baselines present trait cues directly in natural language, which the models appear to process more fluently in this context. This mechanism is consistent with the observed pattern that Likert underperforms while language-native inputs yield cleaner, more stable behavior mappings.

Finally, across models and vignettes, BFI-2-Expanded generally attained the strongest behavioral prediction. We interpret this advantage as the result of Expanded prompts providing richer, sentence-level semantics that minimize representational indirection and better leverage language-native processing. The binary baselines, while more intuitive than Likert, sacrifice granularity; the Likert format, while granular, imposes a numeric-to-semantic translation burden. Taken together, these results suggest that, in this setting, expressing trait information as full sentences (Expanded) offers the most effective and human-like linkage between measured traits and predicted behaviors.

\begin{table}[!htbp]
\caption{Standardized Coefficients For Predicting the Risk-Taking and Moral Dilemma Vignettes}
\label{tab:combined_coefficients}
\small
\centering
\begin{adjustbox}{max width=\textwidth, max totalheight=\textheight, keepaspectratio}
\begin{tabular}{lll *{5}{S}}
\toprule
\multicolumn{3}{c}{} & \multicolumn{5}{c}{\textbf{Predictor}}\\
\cmidrule(l){4-8}
\textbf{Vignette} & \textbf{Format} & \textbf{Model} & \textbf{O} & \textbf{C} & \textbf{E} & \textbf{A} & \textbf{N}\\

\midrule
\multirow{21}{*}{Risk} &  & human     & .14* & -.04 & .24*** & -.03 & -.19** \\
\addlinespace
& Simple-Binary    & GPT-3.5 & .34*** & .06 & .51*** & .14* & -.15* \\
&                   & GPT-4 & .32*** & .15* & .49*** & .20*** & -.36*** \\
&                   & GPT-4o & .41*** & .09 & .52*** & .17** & -.21*** \\
&                   & Llama & .27*** & .03 & .46*** & .13* & -.30*** \\
&                   & Deepseek & .28*** & .11 & .48*** & .14* & -.32*** \\
\addlinespace
& Elaborated-Binary    & GPT-3.5 & .31*** & .08 & .52*** & .19** & -.16** \\
&                   & GPT-4 & .32*** & .15* & .52*** & .22*** & -.30*** \\
&                   & GPT-4o & .39*** & .07 & .50*** & .17** & -.21*** \\
&                   & Llama & .31*** & .06 & .50*** & .18** & -.31*** \\
&                   & Deepseek & .26*** & .16** & .42*** & .18** & -.34*** \\
\addlinespace
& BFI-2-Likert    & GPT-3.5 & .13* & .23*** & .11 & .06 & -.13* \\
&                   & GPT-4 & -.05 & -.12* & .08 & -.13* & -.01 \\
&                   & GPT-4o & -.07 & -.08 & .10 & -.16** & -.13* \\
&                   & Llama & .08 & .17** & .31*** & -.08 & -.20*** \\
&                   & Deepseek & .08 & .07 & .02 & -.19*** & -.05 \\
\addlinespace
& BFI-2-Expanded    & GPT-3.5 & .19** & .02 & .65*** & .13* & -.29*** \\
&                   & GPT-4 & .23*** & .07 & .76*** & .17** & -.35*** \\
&                   & GPT-4o & .32*** & .14* & .81*** & .19** & -.41*** \\
&                   & Llama & .35*** & .03 & .73*** & -.08 & -.24*** \\
&                   & Deepseek & .25*** & .09 & .69*** & .02 & -.36*** \\
\addlinespace

\midrule

\multirow{21}{*}{Moral} &  & human     & -.03 & .21*** & .01 & .05 & -.15* \\
\addlinespace
& Simple-Binary    & GPT-3.5 & -.03 & .10 & .15* & .10 & -.11 \\
&                   & GPT-4 & -.02 & .24*** & .15* & .07 & -.31*** \\
&                   & GPT-4o & -.12* & .35*** & -.06 & .03 & -.25*** \\
&                   & Llama & -.14* & .11 & -.03 & .08 & -.23*** \\
&                   & Deepseek & -.02 & .13* & -.06 & .03 & -.04 \\
\addlinespace
& Elaborated-Binary    & GPT-3.5 & .02 & .04 & .17** & .09 & -.14* \\
&                   & GPT-4 & -.06 & .11 & -.02 & .06 & -.17** \\
&                   & GPT-4o & -.15* & .31*** & -.04 & .04 & -.32*** \\
&                   & Llama & -.16** & .18** & .07 & .08 & -.21*** \\
&                   & Deepseek & -.09 & .27*** & .05 & .02 & -.25*** \\
\addlinespace
& BFI-2-Likert    & GPT-3.5 & .04 & .18** & .17** & .00 & -.01 \\
&                   & GPT-4 & .07 & .23*** & .14* & .16** & -.08 \\
&                   & GPT-4o & -.11 & -.13* & -.21*** & -.31*** & .22*** \\
&                   & Llama & .00 & -.02 & -.07 & -.07 & .00 \\
&                   & Deepseek & .03 & .17** & -.08 & .17** & .04 \\
\addlinespace
& BFI-2-Expanded    & GPT-3.5 & -.06 & .16** & .02 & -.20*** & -.08 \\
&                   & GPT-4 & .01 & .24*** & -.07 & -.04 & -.10 \\
&                   & GPT-4o & .04 & .57*** & -.03 & .02 & -.31*** \\
&                   & Llama & -.04 & .51*** & .06 & .12* & -.28*** \\
&                   & Deepseek & -.05 & .38*** & .04 & .07 & -.23*** \\
\addlinespace
\bottomrule
\end{tabular}
\end{adjustbox}

\footnotesize
\par\smallskip
\raggedright
\textit{Note.} Entries are standardized coefficients (two decimals; leading zero omitted). $^{*}p<.05$, $^{**}p<.01$, $^{***}p<.001$. O = Openness, C = Conscientiousness, E = Extraversion, A = Agreeableness, N = Neuroticism.

\end{table}

\section{General Discussion}
The present research introduces a novel methodology for assigning quantifiable, controllable, and psychometrically validated personalities to Agents using the Big Five personality framework. 
Through a series of three studies, we  demonstrated both the potential and the limitations of this approach for social science research.

Study 1 demonstrated semantic similarities between different Big Five personality measures within the embedding space of LLMs. This finding suggests that LLMs are capable of interpreting and representing personality-related concepts, providing the basis for subsequent studies aimed at validating AI-Agents with assigned Big Five personalities.

In Study 2, we designed a pipeline to create AI-Agents with personalities using prompts in the BFI-2 Likert and Expanded formats. We then validate these AI-Agents by examining their alignment with human participants’ responses on a criterion measure, the Mini-Markers test, and compare them to AI-Agents created with binary adjective prompts (i.e., the Simple- and Elaborated-Binary conditions). Our results show that compared to AI-Agents created with binary adjective prompts or old LLM (i.e., GPT-3.5),  AI-Agents created with the BFI-2-Likert and -Expanded prompts and with newer and more capable LLMs (i.e., GPT-4, GPT-4o, Llama, and DeepSeek) are much more aligned with human responses in terms of item means, item distributions, factor structure, reliability, and correlations between the BFI-2 input and the Mini-Markers outputs. However, this alignment is far from perfect. For example, although the average factor loading magnitudes were similar between AI-Agents (in BFI-2-Likert and -Expanded conditions with newer LLMs) and humans, the specific patterns of loading sizes (e.g., which items had the highest loadings) are quite different.

In Study 3, we further validated the AI-Agents with real-life decision-making vignettes describing risk-taking situations and moral dilemmas. For predicting decisions from personalities, AI-Agents created with the BFI-2-Expanded prompts achieved the strongest and most human-like behavior prediction, while BFI-2-Likert led to the worst performing AI-Agent. One possible explanation is that natural-language inputs (BFI-2-Expanded) reduce representational indirectness relative to numeric inputs (BFI-2-Likert), suggesting the Expanded prompt has the best potential in creating AI-Agents with realistic profiles. For the item-level response pattern, consistent Study 2, AI-Agents created with the BFI-2-Likert or -Expanded prompts and the newer LLMs outperformed those created with binary adjectives. Nevertheless, as expected, newer LLMs had clearly been fine-tuned for safety, resulting in AI-Agents that consistently produced morally inclined responses to the dilemma vignettes, regardless of their assigned personalities. Interestingly, even though these AI-Agents’ responses were heavily skewed towards the high end of the moral scale (typically 8-10 on a 1-10 scale), correlations between Big Five traits and moral dilemma scores closely mirror those observed in human participants. This pattern of results suggests that while fine-tuning shifted the overall response distribution towards more morally inclined responses, the underlying relationships between variables (e.g., between Big Five traits and moral dilemma responses) remain relatively unaffected. 

Overall, our findings demonstrate both the potential and limitations of using AI-Agents as stand-ins for human participants in psychological research. On the one hand, alignment between AI-Agents and humans, particularly in correlations between input Big Five traits and output responses, suggests that AI-Agents hold strong potential as useful tools for preliminary investigations or pilot studies. On the other hand, the discrepancies we observed, such as skewed responses in the moral dilemma vignettes and divergences in the finer patterns of results, underscore that AI-Agents cannot fully substitute for human participants when drawing inferences in large-scale research projects yet.

\subsection{Limitations and Future Directions}

There are several limitations to our study. One limitation is that our method of simulating AI-Agents using only prompts designed from the BFI-2 falls short of capturing the intricate interplay of other important personality constructs, as well as individual backgrounds and societal complexity. 
We acknowledge that personality traits, while central to much of personality psychology, represent only one facet of what most people consider to be ``personality.'' Traits capture relatively stable patterns of thinking, feeling, and behaving, but they do not fully encompass other key components such as values, goals, roles, social identity, and emotional dynamics.  Recent work has begun to investigate these other aspects in LLMs, such as how LLMs represent emotions \parencite{li2023large}, human values \parencite{yao2025value}, or identity \parencite{wang2025large}. These studies highlight that personality in artificial agents can, in principle, extend far beyond trait-level description. Building on these studies, a more integrative approach to LLM personality simulation could combine trait-based control with frameworks for modeling values, motivational systems, and identity-related processes. Such an approach would allow researchers to capture not only how an agent behaves across contexts, but also why it behaves that way, and how it might adapt in response to changing goals or social roles. Expanding in this direction could make AI-Agents a richer platform for studying complex aspects of human personality, bridging dispositional, motivational, and narrative analyses.

Another important limitation concerns the distinction between genuine personality expression and role-playing in AI-Agents. We assigned personality traits to AI-Agents and instructed them to enact these roles using validated psychometric tools. This design choice avoids claims that AI-Agents possess intrinsic or genuine personalities; instead, we leverage their capacity to simulate assigned traits with fidelity and experimental control.
We acknowledge that this role-playing framework may limit generalizability. The AI-Agents’ trait expression is context-dependent, shaped by the prompts and scenarios provided, and may not extend beyond those settings.  Our primary focus is on the methodological utility of AI-Agents as controllable and transparent tools for simulating personality variation in structured research contexts. We do not make claims about consciousness, subjective experience, or ``humanhood'' in AI, which are questions that remain open for philosophical and interdisciplinary debate. Instead, we position our work as a practical contribution to personality and behavioral research, while acknowledging its limitations and the broader questions that lie beyond its scope. 

Finally, due to the large number of conditions in our existing studies, we did not examine LLM parameters such as temperature and top-p; these were kept at their default values across our studies. Theoretically, these parameters could be optimized to generate responses that more closely align with human data. We are currently addressing this in a follow-up study on AI-Agents, using machine learning techniques to optimize these settings and improve response quality.

\subsection{Conclusion}
In conclusion, this research presents a novel and promising approach to creating AI-Agents with psychometrically valid personality traits. Through a series of studies, we showed that AI-Agents created with newer LLMs and BFI-2 prompts have the potential to be used as stand-ins for primary investigations and pilot studies, especially for examining the relationships between personality-related variables. 
However, AI-Agents' responses can differ from those of human participants in many finer patterns of results and in moral-dilemma vignettes, and thus they cannot replace humans in full-scale research yet. Future work may extend this approach by incorporating additional personality variables and demographic factors to capture the more nuanced aspects of human personality.

\section{Acknowledgement}
We gratefully acknowledge Dr. Oliver John for providing original Big Five data crucial for Study 2. We also thank Microsoft Accelerating Foundation Models Research (AFMR) Program and OpenAI's Researcher Access Program for providing API access and funding support for computational resources used in this research. Finally, we thank the editor and reviewers for providing us with suggestions that greatly improved the quality of our manuscript. 



\newpage
\printbibliography


\newpage
\appendix
\renewcommand{\thepage}{A\arabic{page}} 
\setcounter{page}{1} 

\label{sec:appendix_A}
\raggedright
\textbf{Appendix I: Prompts for the Mini-Markers test following personality assignment}

\#\#\# Objective \#\#\#\newline 
Fill out a personality questionnaire. Your questionnaire answers should be reflective of your assigned personalities.\newline 

\#\#\# Response Format \#\#\#\newline 
ONLY return your response as a JSON file where the keys are the traits and the numbers indicate your endorsement to the statements.\newline 

\#\#\# Questionnaire Instruction \#\#\#\newline 
I will provide you a list of descriptive traits. For each trait, take a deep breath and think about what personality you are assigned with then, choose a number indicating how accurately that trait describes you. Using the following rating scale:\newline 
1 - Extremely Inaccurate \newline 
2 - Very Inaccurate\newline 
3 - Moderately Inaccurate \newline 
4 - Slightly Inaccurate \newline 
5 - Neutral / Not Applicable \newline 
6 - Slightly Accurate \newline 
7 - Moderately Accurate \newline 
8 - Very Accurate \newline 
9 - Extremely Accurate \newline  \newline 
\#\#\#  Questionnaire Item \#\#\# \newline 
1. Bashful \_ \newline 
2. Bold \_ \newline 
3. Careless \_ \newline 
4. Cold \_ \newline 
5. Complex \_ \newline 
6. Cooperative \_ \newline 
7. Creative \_ \newline 
8. Deep \_ \newline 
9. Disorganized \_ \newline 
10. Efficient \_ \newline 
11. Energetic \_ \newline 
12. Envious \_ \newline 
13. Extraverted \_ \newline 
14. Fretful \_ \newline 
15. Harsh \_ \newline 
16. Imaginative \_ \newline 
17. Inefficient \_ \newline 
18. Intellectual \_ \newline 
19. Jealous \_ \newline 
20. Kind \_ \newline 
21. Moody \_ \newline 
22. Organized \_ \newline 
23. Philosophical \_ \newline 
24. Practical \_ \newline 
25. Quiet \_ \newline 
26. Relaxed \_ \newline 
27. Rude \_ \newline 
28. Shy \_ \newline 
29. Sloppy \_ \newline 
30. Sympathetic \_ \newline 
31. Systematic \_ \newline 
32. Talkative \_ \newline 
33. Temperamental \_ \newline 
34. Touchy \_ \newline 
35. Uncreative \_ \newline 
36. Unenvious \_ \newline 
37. Unintellectual \_ \newline 
38. Unsympathetic \_ \newline 
39. Warm \_ \newline 
40. Withdrawn \_ \newline  

\newpage
\newpage
\label{sec:appendix_C}

\textbf{Appendix II: Prompts for the risk-taking vignettes following personality assignment}
\raggedright

\#\#\#  Background \#\#\# \newline 
Your decisions and actions are impacted by your personality.\newline  \newline  
\#\#\#  Objective \#\#\# \newline 
I will ask you a sequence of 5 scenario-based questions. Please choose a value between 1 and 10 to indicate your likelihood of an action\newline  \newline  
\#\#\#  Response Format \#\#\# \newline 
Respond with a JSON file in the format of {scenario\_name: likelihood\_value}.\newline  \newline  
\#\#\#  Background \#\#\# \newline 
\#\#\#  Scenario 1 \#\#\# \newline 
Scenario Name: Entrepreneurial\_Venture \newline 
You've developed an innovative business idea that you believe has the potential to be highly successful. Pursuing this would require you to quit your stable job and invest a substantial portion of your savings into the venture. While the idea could revolutionize the industry, there's also a significant risk of failure, which could leave you without a job and deplete your savings. On a scale of 1 to 10, where 1 means pursuing the entrepreneurial venture with all your resources, embracing the risk for the chance of substantial reward, and 10 means conserving your current job and financial security, how would you rate your likely action?\newline  \newline  
\#\#\#  Scenario 2 \#\#\# \newline 
Scenario Name: Investment\newline 
You have recently come into a modest sum of money and are contemplating how to invest it. You're torn between a high-risk, high-reward stock that has the potential to either significantly increase your wealth or decrease in value, and a low-risk government bond that offers guaranteed but modest returns. On a scale of 1 to 10, where 1 means fully embracing the potential high rewards of the risky stock, and 10 means preferring the safety and guaranteed return of the government bond, how would you rate your likely action? \newline  \newline  
\#\#\#  Scenario 3 \#\#\# \newline 
Scenario Name: Confessing\_Feelings \newline 
You find yourself developing romantic feelings for your best friend. These feelings have grown stronger over time, and you're contemplating whether to confess. Doing so could potentially lead to a deeper, romantic relationship, enriching your connection. However, there's also the risk that your confession could make things awkward and possibly harm the friendship you deeply value. On a scale of 1 to 10, where 1 means opening up about your feelings, risking the friendship for the possibility of something more, and 10 means preserving the current friendship without risking discomfort or loss, how would you rate your likely action? \newline  \newline  
\#\#\#  Scenario 4 \#\#\# \newline 
Scenario Name: Extreme\_Sports \newline 
You've always been fascinated by extreme sports and are considering taking up base jumping, drawn by the thrill and the challenge it presents. However, you're fully aware of the significant physical risks involved, including the potential for severe injury. On a scale of 1 to 10, where 1 means diving into the sport for the ultimate thrill despite the risks, and 10 means prioritizing your physical safety and choosing not to participate, how would you rate your likely action? \newline  \newline  
\#\#\#  Scenario 5 \#\#\# \newline  
Scenario Name: Study\_Overseas \newline  
You have been offered admission to a prestigious university in a foreign country, an opportunity that promises to significantly advance your career and personal development. This educational pursuit, however, comes with a substantial sacrifice: you would need to leave behind your family, friends, and the comfort of your familiar environment. The move entails not only physical relocation but also adapting to a new culture, potentially facing language barriers, and starting anew without your established support network. On a scale of 1 to 10, where 1 means embracing the opportunity to study abroad, fully committing to the personal and professional growth it offers despite the sacrifices, and 10 means choosing to stay in your home country to maintain your current relationships and stability, how would you rate your likely action?

\newpage
\label{sec:appendix_D}
\textbf{Appendix III: Prompts for the moral dilemma vignettes following personality assignment}
\raggedright

\#\#\#  Background \#\#\# \newline 
Your decisions and actions are impacted by your personality.\newline  \newline  
\#\#\#  Objective \#\#\# \newline 
I will ask you a sequence of 5 scenario-based questions. Please choose a value between 1 and 10 to indicate your likelihood of an action\newline  \newline  
\#\#\#  Response Format \#\#\# \newline 
Respond with a JSON file in the format of {scenario\_name: likelihood\_value}.\newline  \newline  
\#\#\#  Background \#\#\# \newline 
\#\#\#  Scenario 1 \#\#\# \newline 
Scenario Name: Exam\_Cheating\newline 
You are a student in a community college. During the final exams, you notice your best friend, who has been under a lot of personal stress lately, subtly looking at cheat notes. This exam is crucial for their scholarship. Considering your principles and the empathy you feel for your friend's situation, on a scale of 1 to 10, where 1 means being completely sympathetic and helping your friend out, and 10 means strictly adhering to legal and ethical standards by reporting the cheating, how would you rate your likely action?\newline  \newline  
\#\#\#  Scenario 2 \#\#\# \newline 
Scenario Name: Underage\_Drinking\newline 
At a family reunion, your teenage nephew asks if you can buy alcohol for a party he and his friends are planning. You know he's underage and that supplying alcohol would be illegal, but you also understand his desire to fit in. On a scale of 1 to 10, where 1 means showing understanding and finding a way to support your nephew's request, and 10 means strictly adhering to the law and ethical standards by refusing, how would you rate your likely action?\newline  \newline  
\#\#\#  Scenario 3 \#\#\# \newline 
Scenario Name: Workplace\_Theft\newline 
You work in an office and discover that a colleague, a single parent struggling financially, has been taking office supplies home for their children's school projects. Understanding their struggle but also the implications of theft, on a scale of 1 to 10, where 1 represents full empathy towards your colleague's situation by possibly ignoring or assisting, and 10 represents strict adherence to company policies by reporting the theft, how would you rate your likely action?\newline  \newline  
\#\#\#  Scenario 4 \#\#\# \newline 
Scenario Name: Confidential\_Info\newline 
While volunteering at a local health clinic, you accidentally overhear confidential information about a neighbor's serious health condition. Sharing this could mobilize community support for them, but it would also breach confidentiality. On a scale of 1 to 10, where 1 means prioritizing empathy by finding a way to indirectly support your neighbor, and 10 means upholding the strictest confidentiality standards despite knowing the potential benefits of community support, how would you rate your likely action?\newline  \newline  
\#\#\#  Scenario 5 \#\#\# \newline 
Scenario Name: Honest\_Feedback\newline 
A colleague and friend is up for a performance review that could affect their career. They've been underperforming and now seek your honest feedback. Being completely honest could harm their career and your relationship. On a scale of 1 to 10, where 1 means being sympathetic and possibly softening your feedback to protect their feelings and career, and 10 means giving brutally honest feedback in adherence to your values of honesty and growth, how would you rate your likely action?

\end{document}